# Probabilistic Relational Planning
# with First Order Decision Diagrams

**Saket Joshi**　　　　　　　　　　　　　　　　　　　　JOSHI@EECS.OREGONSTATE.EDU
*School of Electrical Engineering and Computer Science*
*Oregon State University*
*Corvallis, OR 97331, USA*

**Roni Khardon**　　　　　　　　　　　　　　　　　　　　RONI@CS.TUFTS.EDU
*Department of Computer Science*
*Tufts University*
*Medford, MA, 02155, USA*

## Abstract

Dynamic programming algorithms have been successfully applied to propositional stochastic planning problems by using compact representations, in particular algebraic decision diagrams, to capture domain dynamics and value functions. Work on symbolic dynamic programming lifted these ideas to first order logic using several representation schemes. Recent work introduced a first order variant of decision diagrams (FODD) and developed a value iteration algorithm for this representation. This paper develops several improvements to the FODD algorithm that make the approach practical. These include, new reduction operators that decrease the size of the representation, several speedup techniques, and techniques for value approximation. Incorporating these, the paper presents a planning system, FODD-PLANNER, for solving relational stochastic planning problems. The system is evaluated on several domains, including problems from the recent international planning competition, and shows competitive performance with top ranking systems. This is the first demonstration of feasibility of this approach and it shows that abstraction through compact representation is a promising approach to stochastic planning.

## 1. Introduction

Planning under uncertainty is one of the core problems of Artificial Intelligence. Over the years research on automated planning has produced a number of planning formalisms and systems. The STRIPS planning system (Fikes & Nilsson, 1971) led a generation of automated planning research. This produced a number of successful systems for deterministic planning using various paradigms like partial order planning (Penberthy & Weld, 1992), planning based on planning graphs (Blum & Furst, 1997), planning by satisfiability (Kautz & Selman, 1996) and heuristic search (Bonet & Geffner, 2001). These ideas were later employed in solving the problem of planning under uncertainty (Blum & Langford, 1998; Weld, Anderson, & Smith, 1998; Majercik & Littman, 2003; Yoon, Fern, & Givan, 2007; Teic, Koenigsbuch, Infantes, & Kuter, 2008). Of these, approaches using forward heuristic search related to the planning graph (Blum & Furst, 1997) have been very successful at the recent international planning competitions (Yoon et al., 2007; Teichteil-Koenigsbuch et al., 2008).

Another approach to probabilistic planning is based on Markov decision processes (MDPs). The fact that solutions to MDPs generate policies rather than action sequences is particu-larly attractive for probabilistic planning, and this approach came to be known as Decision Theoretic Planning (Boutilier, Dean, & Hanks, 1999a). Classical solution techniques for MDPs, like value iteration (VI) (Bellman, 1957) and policy iteration (PI) (Howard, 1960), are based on dynamic programming. These early solutions, however, require enumeration of the state space. Owing to the curse of dimensionality (Bellman, 1957), even for reasonably small problems, the state space can be very large. This can be seen easily for propositionally factored domains where the state is defined by $N$ binary variables and the number of possible states is $2^N$.

Several approaches were developed to handle such propositionally factored domains (Boutilier, Dearden, & Goldszmidt, 1999b; Kearns & Koller, 1999; Guestrin, Koller, Parr, & Venkataraman, 2003b; Hoey, St-Aubin, Hu, & Boutilier, 1999). One of the most successful, SPUDD (Hoey et al., 1999), demonstrated that if the MDP can be represented using algebraic decision diagrams (ADDs) (Bahar, Frohm, Gaona, Hachtel, Macii, Pardo, & Somenzi, 1993), then VI can be performed entirely using the ADD representation thereby avoiding the need to enumerate the state space. Propositionally factored representations show an impressive speedup by taking advantage of the propositional domain structure. However, they do not benefit from the structure that exists with objects and relations. Boutilier, Reiter, and Price (2001) developed the foundations for provably optimal solutions of relational problems and provided the Symbolic Dynamic Programming (SDP) algorithm in the context of situation calculus. This algorithm provided a framework for dynamic programming solutions of Relational MDPs that was later employed in several formalisms and systems (Kersting, van Otterlo, & De Raedt, 2004; Hölldobler, Karabaev, & Skvortsova, 2006; Sanner & Boutilier, 2009; Wang, Joshi, & Khardon, 2008).

The advantage of the relational representation is abstraction. One can plan at the abstract level without grounding the domain, potentially leading to more efficient algorithms. In addition, the solution at the abstract level is optimal for every instantiation of the domain and can be reused for multiple problems. However, this approach raises some difficult computational issues because one must use theorem proving to reason at the abstract level, and because for some problems optimal solutions at the abstract level can be infinite in size. Following Boutilier et al. (2001) several abstract versions of the value iteration (VI) algorithm have been developed using different representation schemes. For example, approximate solutions based on linear function approximations have been developed and successfully applied to several problems from the international planning competitions (Sanner & Boutilier, 2009).

An alternative representation is motivated by the success of algebraic decision diagrams in solving propositional MDPs (Hoey et al., 1999; St-Aubin, Hoey, & Boutilier, 2000). Following this work, relational variants of decision diagrams have been defined and used for VI algorithms (Wang et al., 2008; Sanner & Boutilier, 2009). Sanner and Boutilier report on an implementation that did not scale well to yield exact solutions for large problems. Our previous work (Wang et al., 2008) introduced First Order Decision Diagrams (FODD), and developed algorithms and reduction operators for them. However, the FODD representation requires non-trivial operations for reductions (to maintain small diagrams and efficiency) leading to difficulties with implementation and scaling.

This paper develops several algorithmic improvements and extensions to the FODD based solution that make the approach practical.

First, we introduce new reduction operators, named R10 and R11, that decrease the size of the FODD representation. R10 makes a global analysis of the FODD and removes many redundant portions of the diagram simultaneously. R11 works locally and targets a particular redundancy that arises quite often when two FODDs are composed through a binary operation; a procedure that is used repeatedly in the VI algorithm. We prove the soundness of these reductions showing that when they are applied the diagrams maintain their correct value.

Second, we present a novel FODD operation, $sub\text{-}apart(A, B)$ that identifies minimal conditions (in terms of variables) under which one FODD $A$ dominates the value of another FODD $B$. This new operation simultaneously expands the applicability of the R7 reduction (Wang et al., 2008) to cover more situations and simplifies the test for its applicability, that must be implemented in our system. We prove the soundness of this operation showing that when it is applied with R7 the diagrams maintain their correct value.

Third, we present several techniques to speed up the FODD-based planning algorithm. These include a sound simplification of one of the steps in the algorithm and in addition several approximation techniques that can trade-off accuracy for improvements in run time.

Fourth, we extend the system to allow it to handle action costs and universal goals. Incorporating all these ideas the paper presents FODD-PLANNER, a planning system for solving relational stochastic planning problems using FODDs.

Fifth, we perform an experimental evaluation of the FODD-PLANNER system on several domains, including problems from the recent international planning competition (IPC). The experiments demonstrate that the new reductions provide significant speedup of the algorithm and are crucial for its practicality. More importantly they show that the FODD-PLANNER exhibits competitive performance with top ranking systems from the IPC. To our knowledge this is the first application of a pure relational VI algorithm without linear function approximation to problems of this scale. Our results demonstrate that abstraction through compact representation is a promising approach to stochastic planning.

The rest of the paper is organized as follows. Section 2 gives a short introduction to relational MDPs and FODDs. Section 3 presents techniques to speed up the FODD-PLANNER. In section 4 we introduce new operators for removing redundancies in FODDs. Section 5 describes the FODD-PLANNER system and in Section 6 we present the results of experiments on planning domains from the IPC. Section 7 provides additional discussion of related work and Section 8 concludes with a summary and ideas for future work.

## 2. Preliminaries

This section gives an overview of Relational MDPs, First Order Decision Diagrams and the Symbolic Dynamic Programming algorithm.

### 2.1 Relational Markov Decision Processes

A Markov decision process (MDP) is a mathematical model of the interaction between an agent and its environment (Puterman, 1994). Formally a MDP is a 5-tuple $< S, A, T, R, \gamma >$ defining

- A set of fully observable states $S$.

- A set $A$ of actions available to the agent.

- A state transition function $T$ defining the probability $P(s'|s,a)$ of getting to state $s'$ from state $s$ on taking action $a$.

- A reward function $R(s,a)$ defining the immediate reward achieved by the agent for being in state $s$ and taking action $a$. To simplify notation we assume that the reward is independent of $a$ so that $R(s,a) = R(s)$. However the general case can be handled in the same way.

- A discount factor $0 \leq \gamma \leq 1$ that captures the relative value of immediate actions over future actions.

The objective of solving a MDP is to generate a policy that maximizes the agent's total, expected, discounted, reward. Intuitively, the expected utility or value of a state is equal to the reward obtained in the state plus the discounted value of the state reached by the best action in the state. This is captured by the Bellman equation as $V(s) = Max_a[R(s) + \gamma \Sigma_{s'} P(s'|s,a)V(s')]$. The discount factor $\gamma < 1$ guarantees that $V(s)$ is finite even when considering an infinite number of steps. For episodic tasks such as planning it provides an incentive to find short solutions. The VI algorithm treats the Bellman equation as an update rule $V(s) \leftarrow Max_a[R(s) + \gamma \Sigma_{s'} P(s'|s,a)V(s')]$, and iteratively updates the value of every state until convergence. Once the optimal value function is known, a policy can be generated by assigning to each state the action that maximizes expected value.

A Relational MDP (RMDP) is a MDP where the world is represented by objects and relations among them. A RMDP is specified by

1. A set of world predicates. Each literal, formed by instantiating a predicate using objects from the domain, can be either `true` or `false` in a given state. For example, in the boxworld domain, world literals are of the form box-in-city($box, city$), box-on-truck($box, truck$), and truck-in-city($truck, city$).

2. A set of action predicates. Each action literal formed by instantiating an action predicate using objects from the domain defines a concrete action. In the boxworld domain, actions have the form load-box-on-to-truck-in-city($box, truck, city$), unload-box -from-truck-in-city($box, truck, city$), and drive-truck($truck, source.city, dest.city$).

3. A state transition function that provides an abstract description of the probabilistic move from one state to another. For example, using a STRIPS-like notation, the transition defined by the action load-box-on-to-truck-in-city can be described as

   **Action:** load-box-on-to-truck-in-city($box, truck, city$):
   **Preconditions:** box-in-city($box, city$), truck-in-city($truck, city$)
   **Outcome 1:** *Probability* 0.8 box-on-truck($box, truck$), ¬box-in-city($box, city$)
   **Outcome 2:** *Probability* 0.2 nothing changes.

   If the preconditions of the action, box-in-city($box, city$) and truck-in-city($truck, city$) are satisfied, then with probability 0.8, the action will succeed generating the effect box-on-truck($box, truck$) and ¬box-in-city($box, city$). All other predicate instantiations remain unchanged. The state remains unchanged with probability 0.2.

4. An abstract reward function describing conditions under which rewards are obtained. For example in the boxworld domain, the reward function is described as $\forall box \forall city$, $destination(box, city) \rightarrow$ box-in-city$(box, city)$ constructed so as to capture the goal of transporting all boxes from their source cities to their respective destination cities.

Boutilier et al. (2001) developed SDP, the first VI algorithm for RMDPs. This was an important theoretical result for RMDPs because for a finite horizon, SDP is guaranteed to produce the optimal value function independent of the domain size. Thus the same value function is applicable for a logistics problem with 2 cities, 2 trucks and 2 boxes, a logistics problem with 100 cities, 1000 trucks and 2000 boxes, and any other instance of the domain.

One of the important ideas in SDP was to represent stochastic actions as deterministic alternatives under nature's control. This helps separate regression over deterministic action alternatives from the probabilities of action effects. This separation is necessary when transition functions are represented as relational schemas abstracting over the structure of the states. The basic outline of the relational value iteration algorithm is as follows:

1. **Regression:** The $n$ step-to-go value function $V_n$ is regressed over every deterministic variant $A_j(\vec{x})$ of every action $A(\vec{x})$ to produce $Regr(V_n, A_j(\vec{x}))$. At the first iteration $V_0$ is assigned the reward function. This is not necessary for correctness of the algorithm but is a convenient starting point for VI. $Regr(V_n, A_j(\vec{x}))$ describes the conditions under which the action alternative $A_j(\vec{x})$ causes the state to transition to some abstract state description in $V^{n+1}$.

2. **Add Action Variants:** The Q-function

$$Q_{V_n}^{A(\vec{x})} = R \oplus [\gamma \otimes \oplus_j (prob(A_j(\vec{x})) \otimes Regr(V_n, A_j(\vec{x})))]$$

for each action $A(\vec{x})$ is generated. In this step the different alternatives of an action are combined. Each alternative $A_j(\vec{x})$ produces $Regr(V_n, A_j(\vec{x}))$ from the regression step. All the $Regr(V_n, A_j(\vec{x}))$s are then added each weighted by the probability of $A_j(\vec{x})$. This produces the parametrized function $Q_{V_n}^{A(\vec{x})}$ which describes the utility of being in a state and taking a concrete action $A(\vec{x})$.

3. **Object Maximization:** Maximize over the action parameters of $Q_{V_n}^{A(\vec{x})}$ to produce $Q_{V_n}^A$ for each action $A(\vec{x})$, thus obtaining the value achievable by the best ground instantiation of $A(\vec{x})$.

4. **Maximize over Actions:** The $n + 1$ step-to-go value function $V_{n+1} = \max_A Q_{V_n}^A$, is generated.

In this description of the algorithm all intermediate constructs ($R$, $P$, $V$ etc.) are represented in some compact form and they capture a mapping from states to values or probabilities. The operations of the Bellman update are performed over these functions while maintaining the compact form. The variant of SDP developed in our previous work (Wang et al., 2008) employed First Order Decision Diagrams to represent the intermediate constructs.

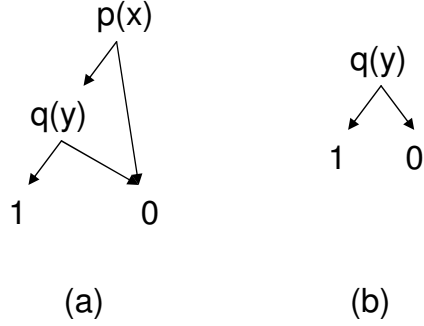

Figure 1: Two example FODDs. In these and all diagrams in the paper, left going edges represent the branch taken when the predicate is `true` and right going edges represent `false` branches.

## 2.2 First Order Decision Diagrams

This section briefly reviews previous work on FODDs and their use for relational MDPs (Wang et al., 2008). We use standard terminology from First-Order logic (Lloyd, 1987). A First Order Decision Diagram is a labeled directed acyclic graph, where each non-leaf node has exactly 2 outgoing edges with `true` and `false` labels. The non-leaf nodes are labeled by atoms generated from a predetermined signature of predicates, constants and an enumerable set of variables. Leaf nodes have non-negative numeric values. The signature also defines a total order on atoms, and the FODD is ordered with every parent smaller than the child according to that order. Two examples of FODDs are given in Figure 1; in these and all diagrams in the paper left going edges represent the `true` branches and right edges are the `false` branches.

Thus, a FODD is similar to a formula in first order logic in that it shares some syntactic elements. Its meaning is similarly defined relative to interpretations of the symbols. An *interpretation* defines a domain of objects, identifies each constant with an object, and specifies a truth value of each predicate over these objects. In the context of relational MDPs, an interpretation represents a state of the world with the objects and relations among them. Given a FODD and an interpretation, a *valuation* assigns each variable in the FODD to an object in the interpretation. Following Groote and Tveretina (2003), the semantics of FODDs are defined as follows. If $B$ is a FODD and $I$ is an interpretation, a valuation $\zeta$ that assigns a domain element of $I$ to each variable in $B$ fixes the truth value of every node atom in $B$ under $I$. The FODD $B$ can then be traversed in order to reach a leaf. The value of the leaf is denoted $Map_B(I, \zeta)$. $Map_B(I)$ is then defined as $max_\zeta Map_B(I, \zeta)$, i.e. an aggregation of $Map_B(I, \zeta)$ over all valuations $\zeta$. For example, consider the FODD in Figure 1(a) and the interpretation $I$ with objects $a, b$ and where the only true atoms are $p(a), q(b)$. The valuations $\{x/a, y/a\}$, $\{x/a, y/b\}$, $\{x/b, y/a\}$, and $\{x/b, y/b\}$, will produce the values $0, 1, 0, 0$ respectively. By the $max$ aggregation semantics, $Map_B(I) = max\{0, 1, 0, 0\} = 1$. Thus, this FODD is equivalent to the formula $\exists x, \exists y, p(x) \wedge q(y)$.

In general, *max* aggregation yields existential quantification when leaves are binary. When using numerical values we can similarly capture value functions for relational MDPs. Thus, every FODD with binary leaves has an equivalent formula in First-Order logic, where all variables are existentially quantified. Conversely, every function free formula in First-Order logic, where the variables are existentially quantified, has an equivalent FODD representation.[1] FODDs cannot capture universal quantification. Recently we introduced a generalized FODD based formalism that does capture arbitrary quantifiers (Joshi, Kersting, & Khardon, 2009); however it is more expensive to use computationally and it is not used in this paper.

Akin to ADDs, FODDs can be combined under arithmetic operations, and reduced in order to remove redundancies. Intuitively, redundancies in FODDs arise in two different ways. The first, observes that some edges will never be traversed by any valuation. Reduction operators for such redundancies are called *strong reduction operators*. The second requires more subtle analysis: there may be parts of the FODD that are traversed under some valuations but because of the max aggregation, the valuations that traverse those parts are never instrumental in determining the map. Operators for such redundancies are called *weak reductions operators*. Strong reductions preserve $Map_B(I, \zeta)$ for every valuation $\zeta$ (thereby preserving $Map_B(I)$) and weak reductions preserve $Map_B(I)$ but not necessarily $Map_B(I, \zeta)$ for every $\zeta$. Groote and Tveretina (2003) introduced four strong reduction operators (R1 $\cdots$ R4). Wang et al. (2008) added the strong reduction operator $R5$. They also introduced the notion of weak reductions and developed weak reduction operators (R6 $\cdots$ R9). Another subtlety arises because for RMDP domains we may have some background knowledge about the predicates in the domain. For example, in the blocksworld, if block $a$ is clear then $on(x, a)$ is false for all values of $x$. We denote such *background knowledge* by $\mathcal{B}$ and allow reductions to rely on such knowledge. Below, we discuss the operator R7 in some detail because of its relevance to the next section.

We use the following notation. If $e$ is an edge from node $n$ to node $m$, then source($e$) = $n$, target($e$) = $m$ and sibling($e$) is the other edge out of $n$. For node $n$, the symbols $n_{\downarrow t}$ and $n_{\downarrow f}$ denote the `true` and `false` edges out of $n$ respectively. $l(n)$ denotes the atom associated with node $n$. Node formulas (NF) and edge formulas (EF) are defined recursively as follows. For a node $n$ labeled $l(n)$ with incoming edges $e_1, \ldots, e_k$, the node formula is NF($n$) = ($\vee_i$EF($e_i$)). The edge formula for the `true` outgoing edge of $n$ is EF($n_{\downarrow t}$) = NF($n$) $\wedge$ $l(n)$. The edge formula for the `false` outgoing edge of $n$ is EF($n_{\downarrow f}$) = NF($n$) $\wedge$ $\neg l(n)$. These formulas, where all variables are existentially quantified, capture the conditions under which a node or edge are reached. Similarly, if $B$ is a FODD and $p$ is a path from the root to a leaf in $B$, then the path formula for $p$, denoted by PF($p$) is the conjunction of literals along $p$. The variables of $p$, are denoted $\vec{x^p}$. When $\vec{x^p}$ are existentially quantified, satisfiability of PF($p$) under an interpretation $I$ is a necessary and sufficient condition for the path $p$ to be traversed by some valuation under $I$. If $\zeta$ is such a valuation, then we define $Path_B(I, \zeta) = p$. The leaf reached by path $p$ is denoted as $leaf(p)$. We let PF($p$)\$Lit$ denote the path formula of path $p$ with the literal $Lit$ removed (if it was present) from the conjunction. $\mathcal{B}$ denotes background knowledge of the domain.

---

1. This can be seen by translating the formula $f$ into a disjunctive normal form $f = \vee f_i$, representing every conjunct $f_i$ as a FODD, and calculating their disjunction using the apply procedure of Wang et al. (2008).

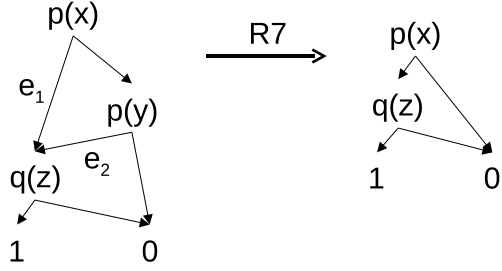

Figure 2: An example of the R7 reduction.

In the process of the algorithm, and also during reductions, we need to perform operations on functions represented as FODDs. Let $B_1$ and $B_2$ be two FODDs each representing a function from states to real values ($B_1 : S \to \Re$, $B_2 : S \to \Re$). Let $B$ be the function such that $\forall S$, $B(S) = B_1(S) + B_2(S)$. Wang et al. (2008) provide an algorithm for calculating a FODD representation of $B$. We denote this operation by $B = B_1 \oplus B_2$ and similarly use $\ominus$, $\otimes$ etc. to denote operations over diagrams.

**The R7 Reduction:** Weak reductions arise in two forms - edge redundancies and node redundancies. Corresponding to these, the R7 reduction operator (Wang et al., 2008) has two variants - R7-replace (for removing redundant edges) and R7-drop (for removing redundant nodes). An edge is redundant when all valuations going through it are dominated by other valuations. Intuitively, given a FODD $B$ and edges $e_1$ and $e_2$ in $B$, if for every valuation going through edge $e_2$, there always is another valuation going through $e_1$ that gives a better value, we can replace target($e_2$) by 0 without affecting $Map_B(I)$ for any interpretation $I$. Figure 2 shows an example of this reduction. In the FODD on the left, consider a valuation reaching the 1 leaf by traversing the path $\neg p(x) \wedge p(y)$ under some interpretation $I$. Then we can generate another valuation (by substituting the value of $y$ for the value of $x$) that reaches the 1 leaf through the path $p(x)$. Therefore, intuitively the path $\neg p(x) \wedge p(y)$ is redundant and can be removed from the diagram. The R7-replace reduction formalizes this notion with a number of conditions such that when certain combinations of these conditions are satisfied, an edge reduction becomes applicable. For example, when the following conditions occur together in a FODD, it can be reduced by replacing the target of edge $e_2$ by the 0 leaf.

**(P7.2)** : $\mathcal{B} \models \forall \vec{u}, [[\exists \vec{w}, \mathrm{EF}(e_2)] \to [\exists \vec{v}, \mathrm{EF}(e_1)]]$ where $\vec{u}$ are the variables that appear in both target($e_1$) and target($e_2$), $\vec{v}$ the variables that appear in $\mathrm{EF}(e_1)$ but are not in $\vec{u}$, and $\vec{w}$ the variables that appear in $\mathrm{EF}(e_2)$ but are not in $\vec{u}$.

This condition requires that for every valuation $\zeta_1$ that reaches $e_2$ there is a valuation $\zeta_2$ that reaches $e_1$ such that $\zeta_1$ and $\zeta_2$ agree on all variables that appear in both $target(e_1)$ and $target(e_2)$.

**(V7.3)** : all leaves in $D = target(e_1) \ominus target(e_2)$ have non-negative values, denoted as $D \geq 0$. In this case for any fixed valuation potentially reaching both $e_1$ and $e_2$ it is better to follow $e_1$ instead of $e_2$.

**(S1)** : There is no path from the root to a leaf that contains both $e_1$ and $e_2$.

The operator R7-replace$(e_1, e_2)$ replaces $target(e_2)$ with a leaf valued 0. Notice that the FODD in Figure 2 satisfies conditions P7.2, V7.3, and S1. For **(P7.2)** the shared variable is $z$ and it holds that $\forall z, [[\exists x \exists y, \neg p(x) \wedge p(y)] \rightarrow [\exists x, p(x)]]$. **(V7.3)** holds because $target(e_1) = target(e_2)$ and $D \equiv 0$. With these definitions Wang et al. (2008) show that it is safe to perform R7-replace when the conditions P7.2, V7.3, and S1 hold:

**Lemma 1 ((Wang et al., 2008))** *Let B be a FODD, $e_1$ and $e_2$ edges for which conditions P7.2, V7.3, and S1 hold, and $B'$ the result of R7-replace$(e_1, e_2)$, then for any interpretation I we have $MAP_B(I) = MAP_{B'}(I)$.*

Similarly R7-drop formalizes conditions under which nodes can be dropped from the diagram. Several alternative conditions for the applicability of R7 (R7-replace and R7-drop) are given by Wang et al. (2008). This provided a set of alternative conditions for applicability of R7 none of which dominates the others, with the result that effectively one has to check all the conditions when reducing a diagram. The next section shows how the process of applying R7 can be simplified and generalized.

R7 captures the fundamental intuition behind weak reductions and hence is widely applicable. Unfortunately it is also very expensive to run. In practice R7-replace conditions have to be tested for all pairs of edges in the diagram. Each test requires theorem proving with disjunctive First-Order formulas.

## 2.3 VI with FODDs

In previous work (Wang et al., 2008) we showed how to capture the reward function and the dynamics of the domain using FODDs and presented a value iteration algorithm along the lines described in the last section. Reward and value functions are captured directly using FODDs. Domains are captured by FODDs describing the probabilities of action variants $prob(A_j(\vec{a}))$, and by special FODDs, Truth Value Diagrams (TVD), that capture the deterministic effects of each action variant, similar to the successor state axioms used by Boutilier et al. (2001). For every action variant $A_j(\vec{a})$ and each predicate schema $p(\vec{x})$ the TVD $T(A(\vec{a}), p(\vec{x}))$ is a FODD with $\{0, 1\}$ leaves. The TVD gives the truth value of $p(\vec{x})$ in the next state when $A(\vec{a})$ has been performed in the current state. The TVDs therefore capture action preconditions within the FODD structure and $p(\vec{x})$ is the potential effect where the formalism specifies its truth value directly instead of saying whether it changes or not. All operations that are needed in the SDP algorithm (regression, *plus*, *times*, *max*) can be performed by special algorithms combining FODDs. The details of these representations and algorithms were previously described (Wang et al., 2008) and they are not directly needed for the discussion in this paper and thus omitted here.

On the other hand, direct application of these operations will yield large FODDs with redundant structure and therefore, to keep the diagram size manageable, FODDs have to be reduced at every step of the algorithm. Efficient and successful reductions are the key to this procedure. The reductions R1-R9 (Groote & Tveretina, 2003; Wang et al., 2008) provide a first step towards an efficient FODD system. However, they do not cover all possible redundancies and they are expensive to apply in practice. Therefore a direct implementation of these is not sufficient to yield an effective stochastic planner. In the following sections we present new reduction operations and speedup techniques to make VI with FODDs practical.

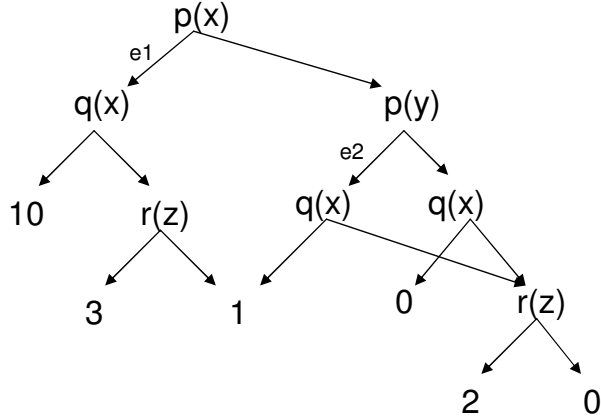

Figure 3: FODD example showing applicability of Sub-Apart.

## 3. Speedup Techniques

This section presents two techniques to speed up the VI algorithm of Wang et al. (2008) while maintaining an exact solution.

### 3.1 Subtracting Apart - Improving Applicability of R7

The applicability of R7 can be increased if certain branches have variables standardized apart in a way that preserves the evaluation of the FODD under the max aggregation semantics. Consider the FODD $B$ in Figure 3. Intuitively a weak reduction is applicable on this diagram because of the following argument. Consider a valuation $\zeta = \{x \setminus 1, y \setminus 2, z \setminus 3\}$ crossing edge $e_2$ under some interpretation $I$. Then $I \models \mathcal{B} \to \neg p(1) \wedge p(2)$. Therefore there must be a valuation $\eta = \{x \setminus 2, z \setminus 3\}$ (and any value for $y$), that crosses edge $e_1$. Now depending on the truth value of $I \models \mathcal{B} \to q(1)$ and $I \models \mathcal{B} \to q(2)$, we have four possibilities of where $\zeta$ and $\eta$ would reach after crossing the nodes target($e_2$) and target($e_1$) respectively. However, in all these cases, $Map_B(I, \eta) \geq Map_B(I, \zeta)$. Therefore we should be able to replace target($e_2$) by a 0 leaf. A similar argument shows that we should also be able to drop the node source($e_2$). Surprisingly, though, none of the R7 conditions apply in this case and this diagram cannot be reduced. On closer inspection we find that the reason for this is that the conditions **(P7.2)** and **(V7.3)** are too restrictive. **(V7.3)** holds but **(P7.2)** requires that $\forall x, \forall z, [[\exists y, \neg p(x) \wedge p(y)] \to [p(x)]]$ implying that for every valuation crossing edge $e_2$, there has to be another valuation crossing edge $e_1$ such that the valuations agree on the value of $x$ and $z$ and this does not hold. However, from our argument above, for $\eta$ to dominate $\zeta$, the two valuations need not agree on the value of $x$. We observe that if we rename variable $x$ so that its instances are different in the sub-FODDs

240

rooted at target($e_1$) and target($e_2$) (i.e. we standardized apart w.r.t. $x$) then both **(P7.2)** and **(V7.3)** go through and the diagram can be reduced. Notice that for this type of simplification to go through it must be the case that $B_1 \ominus B_2 \geq 0$ already holds. The more variables we standardize apart the "harder" it is to keep this condition. To develop this idea, we introduce a new FODD subtraction algorithm *Sub-apart*: Given diagrams $B_1$ and $B_2$ the algorithm tries to standardize apart as many of their common variables as possible, while keeping the condition $B_1 \ominus B_2 \geq 0$ true. The algorithm returns a 2-tuple $\{T, V\}$, where $T$ is a Boolean variable indicating whether the combination can produce a diagram that has no negative leaves when all variables except the ones in $V$ are standardized apart.

The algorithm uses the standard recursive template for combining ADDs and FODDs (Bahar et al., 1993; Wang et al., 2008) where a root node is chosen from the root of the two diagrams and the operation is recursively performed on the corresponding sub-diagrams. In addition when the roots of the two diagrams are identical Sub-apart considers the possibility of making them different by standardizing apart. Sub-apart uses these recursive calls to collect constraints specifying which variables cannot be standardized apart; these sets are combined and returned to the calling procedure.

**Procedure 1** *Sub-apart(A, B)*

1. *If A and B are both leaves,*

    (a) *If $A - B \geq 0$ return $\{true, \{\}\}$ else return $\{false, \{\}\}$*

2. *If $l(A) < l(B)$, let*

    (a) *$\{L, V_1\} = $ Sub-apart(target($A_{\downarrow t}$), B)*

    (b) *$\{R, V_2\} = $ Sub-apart(target($A_{\downarrow f}$), B)*

    *Return $\{L \wedge R, V_1 \cup V_2\}$*

3. *If $l(A) > l(B)$, let*

    (a) *$\{L, V_1\} = $ Sub-apart(A, target($B_{\downarrow t}$))*

    (b) *$\{R, V_2\} = $ Sub-apart(A, target($B_{\downarrow f}$))*

    *Return $\{L \wedge R, V_1 \cup V_2\}$*

4. *If $l(A) = l(B)$, let V be the variables of A (or B). Let*

    (a) *$\{LL, V_3\} = $ Sub-apart(target($A_{\downarrow t}$), target($B_{\downarrow t}$))*

    (b) *$\{RR, V_4\} = $ Sub-apart(target($A_{\downarrow f}$), target($B_{\downarrow f}$))*

    (c) *$\{LR, V_5\} = $ Sub-apart(target($A_{\downarrow t}$), target($B_{\downarrow f}$))*

    (d) *$\{RL, V_6\} = $ Sub-apart(target($A_{\downarrow f}$), target($B_{\downarrow t}$)*

    (e) *If $LL \wedge RR = false$, return $\{false, V_3 \cup V_4\}$*

    (f) *If $LR \wedge RL = false$ return $\{true, V \cup V_3 \cup V_4\}$*

    (g) *Return $\{true, V_3 \cup V_4 \cup V_5 \cup V_6\}$*

The next theorem shows that the procedure is correct. The variables common to $B_1$ and $B_2$ are denoted by $\vec{u}$ and $B^{\vec{w}}$ denotes the combination diagram of $B_1$ and $B_2$ under the subtract operation when all variables except the ones in $\vec{w}$ are standardized apart. Let $n_1$ and $n_2$ be the roots nodes of $B_1$ and $B_2$ respectively.

**Theorem 1** *Sub-apart($n_1$, $n_2$) = {true, $\vec{v}$} implies $B^{\vec{v}}$ contains no negative leaves and Sub-apart($n_1$, $n_2$) = {false, $\vec{v}$} implies $\neg \exists \vec{w}$ such that $\vec{w} \subseteq \vec{u}$ and $B^{\vec{w}}$ contains no negative leaves.*

*Proof:* The proof is by induction on $k$, the sum of the number of nodes in $B_1$ and $B_2$. For the base case when $k = 2$, both $B_1$ and $B_2$ are single leaf diagrams and the statement is trivially true. Assume that the statement is true for all $k \leq m$ and consider the case where $k = m+1$. When $l(n_1) < l(n_2)$, in the resultant diagram of combination under subtraction, we expect $n_1$ to be the root node and $n_{1 \downarrow t} \ominus n_2$ and $n_{1 \downarrow f} \ominus n_2$ to be the left and right sub-FODDs. Hence, the Sub-apart algorithm recursively calls Sub-apart($n_{1 \downarrow t}$, $n_2$) and Sub-apart($n_{1 \downarrow f}$, $n_2$). Since the sum of the number of nodes of the diagrams in the recursive calls is always $\leq m$, the statement is true for both recursive calls. Clearly, the top level can return a `true` iff both calls return `true`. In addition, if we keep the variables in $V_1$ and $V_2$ (of step 2) in their original form (that is, not standardized apart) then for both branches of the new root $n_1$ we are guaranteed positive leaves and therefore the same is true for the diagram rooted at $n_1$. A similar argument shows that the statement is true when $l(n_1) > l(n_2)$.

When $l(n_1) = l(n_2)$, again by the inductive hypothesis, the statement of the theorem is true for all recursive calls. Here we have 2 choices. We could either standardize apart the variables $V$ in $l(n_1)$ and $l(n_2)$ or keep them identical. If they are the same, in the resultant diagram of combination under subtraction we expect $n_1$ to be the root node and $n_{1 \downarrow t} \ominus n_{2 \downarrow t}$ and $n_{1 \downarrow f} \ominus n_{2 \downarrow f}$ to be the left and right sub-FODDs. Again the top level can return a `true` iff both calls return `true`. The set of shared variables requires the variables of $l(n_1)$ in addition to those from the recursive calls in order to ensure that $l(n_1) = l(n_2)$.

If we standardize apart $l(n_1)$ and $l(n_2)$, then we fall back on one of the cases where $n_1 \neq n_2$ except that the algorithm checks for the second level of recursive calls $n_{1 \downarrow t} \ominus n_{2 \downarrow t}$, $n_{1 \downarrow t} \ominus n_{2 \downarrow f}$, $n_{1 \downarrow f} \ominus n_{2 \downarrow t}$ and $n_{1 \downarrow f} \ominus n_{2 \downarrow f}$. The top level of the algorithm can return `true` if all four calls return `true` and return the union of the sets of variables returned by the four calls. If not all four calls return `true`, the algorithm can still keep the variables in $l(n_1)$ and $l(n_2)$ identical and return `true` if the conditions for that case are met. $\square$

The theorem shows that the algorithm is correct but does not guarantee minimality. In fact, the smallest set of variables $\vec{w}$ for $B^{\vec{w}}$ to have no negative leaves may not be unique (Wang et al., 2008). One can also show that the output of Sub-apart may not be minimal. In principle, one can use a greedy procedure that standardizes apart one variable at a time and arrives at a minimal set $\vec{w}$. However, although *Sub-apart* does not produce a minimal set, we prefer it to the greedy approach because it is fast and often generates a small set $\vec{w}$ in practice. We can now define new conditions for applicability of R7:

**(V7.3S)** : Sub-apart(target($e_1$), target($e_2$)) = {$true$, $V_1$}.

**(P7.2S)** : $\mathcal{B} \models \forall V_1, [[\exists \vec{w}, \mathrm{EF}(e_2)] \rightarrow [\exists \vec{v}, \mathrm{EF}(e_1)]]$ where as above and $\vec{v}$, $\vec{w}$ are the remaining variables (i.e. not in $V_1$) in $\mathrm{EF}(e_1)$, $\mathrm{EF}(e_2)$ respectively.

**(P7.2S)** guarantees that whenever there is a $\zeta_2$ running through target($e_2$), there is always a $\zeta_1$ running through target($e_1$) and $\zeta_1$ and $\zeta_2$ agree on $V_1$. **(V7.3S)** guarantees that under this condition, $\zeta_1$ provides a better value than $\zeta_2$. Using exactly the same proof as Lemma 1 given by Wang et al. (2008), we can show the following:

**Lemma 2** *Let $B$ be a FODD, $e_1$ and $e_2$ edges for which conditions P7.2S, V7.3S, and S1 hold, and $B'$ the result of R7-replace($e_1, e_2$), then for any interpretation $I$ we have $MAP_B(I) = MAP_{B'}(I)$.*

Importantly, conditions **(P7.2S)**, **(V7.3S)** subsume all the previous conditions for applicability and safety of R7-replace that were previously given (Wang et al., 2008). Therefore, instead of testing for multiple conditions it is sufficient to test for **(P7.2S)** and **(V7.3S)**. A very similar argument as the one above shows how *Sub-apart* extends and simplifies the conditions of R7-drop. Thus the use of Sub-apart both simplifies the conditions to be tested and provides more opportunities for reductions. In our implementation, we use the new conditions with Sub-apart whenever testing for applicability of R7-replace and R7-drop.

## 3.2 Not Standardizing Apart

Recall that the FODD-based VI algorithm must add functions represented by FODDs (in Steps 2 and 4) and take the maximum over functions represented by FODDs (in Step 4). Since the individual functions are independent functions of the state, the variables of different functions are not related to one another. Therefore, before adding or maximizing, the algorithm by Wang et al. (2008) standardizes apart the diagrams. That is, all variables in the diagrams are given new names so they do not constrain each other. On the other hand, since the different diagrams are structurally related this often introduces redundancies (in the form of renamed copies of the same atoms) that must be removed by reduction operators. However, our reduction operators are not ideal and avoiding this step can lead to significant speedup in the system. Here we observe that for maximization (in Step 4) standardizing apart is not needed and therefore can be avoided.

**Theorem 2** *Let $B_1$ and $B_2$ be FODDs. Let $B$ be the result of combining $B_1$ and $B_2$ under the max operation when $B_1$ and $B_2$ are standardized apart. That is, $\forall s$, $MAP_B(s) = \max\{MAP_{B_1}(s), MAP_{B_2}(s)\}$. Let $B'$ be the result of combining $B_1$ and $B_2$ under the max operation when $B_1$ and $B_2$ are not standardized apart. $\forall$ interpretations $I$, $Map_B(I) = Map_{B'}(I)$.*

*Proof:* The theorem is proved by showing that for any $I$ a valuation for the maximizing diagram can be completed into a valuation over the combined diagram giving the same value. Clearly $Map_B(I) \geq Map_{B'}(I)$ since every substitution and path that exist for $B'$ are also possible for $B$. We show that the other direction holds as well. Let $\vec{u}$ be the variables common to $B_1$ and $B_2$. Let $\vec{w_1}$ be the variables in $B_1$ that are not in $B_2$ and $\vec{w_2}$ be the variables in $B_2$ not in $B_1$. By definition, for any interpretation $I$,

$$Map_B(I) = Max[Map_{B_1}(I), \, Map_{B_2}(I)] = Max[Map_{B_1}(I, \zeta_1), \, Map_{B_2}(I, \zeta_2)]$$

243

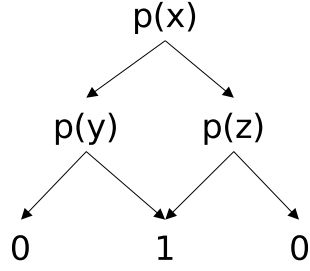

Figure 4: FODD example illustrating the need for a DPO.

for some valuations $\zeta_1$ over $\vec{u}\vec{u_1}$ and $\zeta_2$ over $\vec{u}\vec{u_2}$. Without loss of generality let us assume that $Map_{B_1}(I, \zeta_1) = Max[Map_{B_1}(I, \zeta_1), Map_{B_2}(I, \zeta_2)]$. We can construct valuation $\zeta$ over $\vec{u}\vec{u_1}\vec{u_2}$ such that $\zeta$ and $\zeta_1$ share the values of variables in $\vec{u}$ and $\vec{u_1}$. Obviously $Map_{B_1}(I, \zeta)$ = $Map_{B_1}(I, \zeta_1)$. Also, by the definition of FODD combination, we have $Map_{B'}(I) \geq$ $Map_{B_1}(I, \zeta) = Map_B(I)$. □

## 4. Additional Reduction Operators

In this section we introduce two new reduction operators that improve the efficiency of the VI algorithm. The following definitions are important in developing these reductions and to understand potential scope for reducing diagrams.

**Definition 1** *A descending path ordering (DPO) is an ordered list of all paths from the root to a leaf in FODD B, sorted in descending order by the value of the leaf reached by the path. The relative order of paths reaching the same leaf can be set arbitrarily.*

**Definition 2** *If B is a FODD, and P is the DPO for B, then a path $p_j \in P$ is instrumental with respect to P iff*

1. *there is an interpretation I and valuation, $\zeta$, such that $Path_B(I, \zeta) = p_j$, and*

2. *$\forall$ valuations $\eta$, if $Path_B(I, \eta) = p_k$, then $k \geq j$.*

The example in Figure 4 shows why a DPO is needed. The paths $p(x) \land \neg p(y)$ and $\neg p(x) \land p(z)$ both imply each other. Whenever there is a valuation traversing one of the paths there is always another valuation traversing the other. Removing any one path from the diagram would be safe meaning that the *map* is not changed. But we cannot remove both paths. Without an externally imposed order on the paths, it is not clear which path should be labeled as redundant. A DPO does exactly that to make the reduction possible. It is not clear at the outset how to best choose a DPO so as to maximally reduce the size of a diagram. A lexicographic ordering over paths of equal value makes for an easy implementation but may not be the best. We describe our heuristic approach for choosing DPOs in the next section in the context of the implementation of the FODD-PLANNER.

## 4.1 The R10 Reduction

A path in FODD $B$ is dominated if whenever a valuation traverses it, there is always another valuation traversing another path and reaching a leaf of greater or equal value. Now if all paths through an edge $e$ are dominated, then no valuation crossing that edge will ever determine the map under $max$ aggregation semantics. In such cases we can replace target($e$) by a 0 leaf. This is the basic intuition behind the R10 operation.

Although its objective is the same as that of R7-replace, R10 is faster to compute in some cases and has two advantages over R7-replace. First, because paths can be ranked by the value of the leaf they reach, we can perform a single ranking and check for all dominated paths (and hence all dominated edges). Hence, while all other reduction operators are local, R10 is a global reduction. Second, the theorem proving required for R10 is always on conjunctive formulas with existentially quantified variables, which is decidable in the function free case (e.g., Khardon, 1999). This gives a speedup over R7-replace. On the other hand R10 must explicitly enumerate the DPO and is therefore not efficient if the FODD has an exponential number of non-zero valued paths. In such a case R7 or some other edge based procedure is likely to be more efficient.

Consider the example shown in Figure 5. The following list specifies a DPO for this diagram:

1. $p(y), \neg p(z), \neg p(x) \to 3$

2. $p(y), \neg p(z), p(x), \neg q(x) \to 3$

3. $\neg p(y), p(x), q(x) \to 2$

4. $p(y), \neg p(z), p(x), q(x) \to 2$

Notice that the relative order of paths reaching the same leaf in this DPO is defined by ranking shorter paths higher than longer ones. This is not a requirement for the correctness of the algorithm but is a good heuristic. According to the reduction procedure, all edges of path 1 are important and cannot be reduced. However, since 1 subsumes 2, 3 and 4, all the other edges (those belonging to paths 2, 3 and 4 and those not appearing in any of the ranked paths) can be reduced. Therefore the reduction procedure replaces the targets of all edges other than the ones in path 1, to the value 0. Path 1 is thus an *instrumental* path but paths 2, 3 and 4 are not. This process is formalized in the following algorithm.

**Procedure 2** *R10(B)*

1. *Let $E$ be the set of all edges in $B$*

2. *Let $P = [p_1, p_2 \cdots p_n]$ be a DPO for $B$. Thus $p_1$ is a path reaching the highest leaf and $p_n$ is a path reaching the lowest leaf.*

3. *For $j = 1$ to $n$, do the following*

   (a) *Let $E_{p_j}$ be the set of edges on $p_j$*

   (b) *If $\neg \exists i, \ i < j$ such that $\mathcal{B} \models (\exists x^{\vec{p_j}}, \ PF(p_j)) \to (\exists x^{\vec{p_i}}, \ PF(p_i))$, then set $E = E - E_{p_j}$*

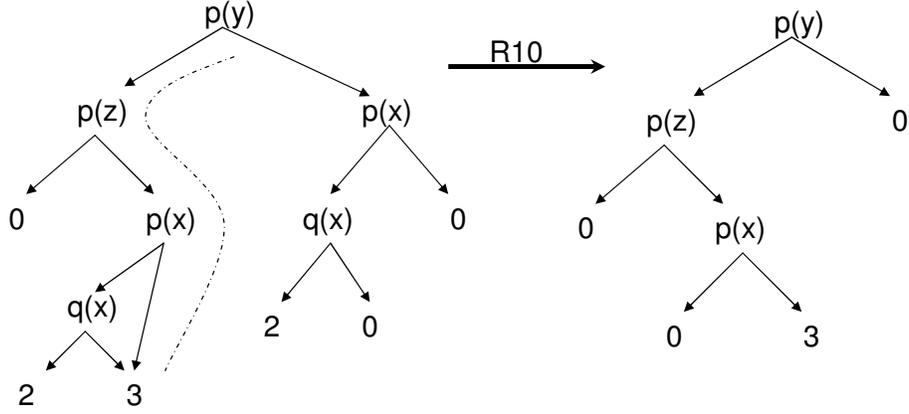

Figure 5: FODD example illustrating the R10 reduction.

*4. For every edge $e \in E$, set target(e) = 0 in B*

In the example in Figure 5 none of the paths 2, 3 and 4 pass the conditions of step 3b in the algorithm. Therefore their edges are not to be removed from $E$ and are assigned the value 0 by the algorithm. Here R10 is able to identify in one pass, the one path (shown along a curved indicator line) that dominates all other paths. To achieve the same reduction, R7-replace takes 2-3 passes depending on the order of application. Since every pass of R7-replace has to check for implication of edge formulas for every pair of edges, this can be expensive. On the other hand, there are cases where R10 is not applicable but R7-replace is. An example of this is shown in the diagram in Figure 6. For this diagram it is easy to see that if $e_2$ is reached then so is $e_1$ and $e_1$ always gives a strictly better value. R10 cannot be applied because it tests subsumption for complete paths. In this case the path for $e_2$ implies the disjunction of two paths going through $e_1$.

We next present a proof of correctness for R10. Lemma 3 shows that our test for instrumental paths is correct. Lemma 4 shows that, as a result, edges marked for deletion at the end of the algorithm do not belong to any instrumental path. The theorem uses this fact to argue correctness of the algorithm.

**Lemma 3** *For any path $p_j \in P$, if $p_j$ is instrumental then $\neg \exists i$, $i < j$ and $\mathcal{B} \models (\exists x \vec{p_j}$, $\mathrm{PF}(p_j)) \rightarrow (\exists x \vec{p_i}$, $\mathrm{PF}(p_i))$.*

*Proof:* If $p_j$ is instrumental then by definition, there is an interpretation $I$ and valuation, $\zeta$, such that $Path_B(I, \zeta) = p_j$, and $\forall$ valuations $\eta$, $\neg \exists i < j$ such that $Path_B(I, \eta) = p_i$. In other words, $I \models [\mathcal{B} \rightarrow (\exists x \vec{p_j}$, $\mathrm{PF}(p_j))]$ but $I \not\models [\mathcal{B} \rightarrow (\exists x \vec{p_i}$, $\mathrm{PF}(p_i))]$ for any $i < j$. This implies that $\neg \exists i$, $i < j$ and $(\mathcal{B} \rightarrow \exists x \vec{p_j}$, $\mathrm{PF}(p_j)) \models (\mathcal{B} \rightarrow \exists x \vec{p_i}$, $\mathrm{PF}(p_i))$. Hence $\neg \exists i$, $i < j$ and $\mathcal{B} \models [(\exists x \vec{p_j}$, $\mathrm{PF}(p_j)) \rightarrow (\exists x \vec{p_i}$, $\mathrm{PF}(p_i))]$. $\square$

**Lemma 4** *If $E$ is the set of edges left at the end of the R10 procedure then if $e \in E$ then there is no instrumental path that goes through $e$.*

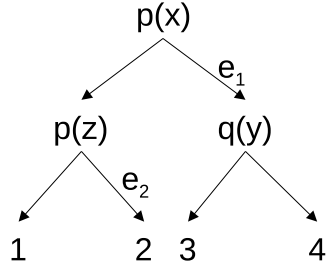

Figure 6: FODD example where R7 is applicable but R10 is not.

*Proof:* Lemma 3 proves that if a path $p_j$ is instrumental, then $\neg \exists i$, $i < j$ and $\mathcal{B} \models [(\exists x \overrightarrow{p_j}, \mathrm{PF}(p_j)) \rightarrow (\exists x \overrightarrow{p_i}, \mathrm{PF}(p_i))]$. Thus in step 3b of R10, if a path is instrumental, all its edges are removed from $E$. Therefore if $e \in E$ at the end of the R10 procedure, it cannot be in $p_j$. Since $p_j$ is not constrained in any way in the argument above, $e$ cannot be in any instrumental path. □

**Theorem 3** *Let $B$ be any FODD. If $B' = R10(B)$ then $\forall$ interpretations $I$, $Map_B(I) = Map_{B'}(I)$.*

*Proof:* By the definition of R10, the only difference between $B$ and $B'$ is that some edges that pointed to sub-FODDs in $B$, point to the 0 leaf in $B'$. These are the edges left in the set $E$ at the end of the R10 procedure. Therefore any valuation crossing these edges achieves a value of 0 in $B'$ but could have achieved more value in $B$ under the same interpretation. Valuations not crossing these edges will achieve the same value in $B'$ as they did in $B$. Therefore for any interpretation $I$ and valuation $\zeta$, $Map_B(I, \zeta) \geq Map_{B'}(I, \zeta)$ and hence $Map_B(I) \geq Map_{B'}(I)$.

Fix any interpretation $I$ and $v = Map_B(I)$. Let $\zeta$ be a valuation such that $Map_B(I, \zeta) = v$. If there is more than one $\zeta$ that gives the value $v$, we choose one whose path $p_j$ has the least index in $P$. Now by definition $p_j$ is instrumental and by lemma 4, none of the edges of $p_j$ are removed by R10. Therefore $Map_{B'}(I, \zeta) = v = Map_B(I)$. Finally, by the definition of the max aggregation semantics, $Map_{B'}(I) \geq Map_{B'}(I, \zeta)$ and therefore $Map_{B'}(I) \geq Map_B(I)$. □

The R10 procedure is similar to the reduction of decision list rules of ReBel (Kersting et al., 2004). The difference, however, is that R10 is a reduction procedure for FODDs and therefore uses the individual rules only as a subroutine to gather information about redundant edges. Thus while ReBel removes paths R10 removes edges affecting multiple paths in the diagram. The main potential disadvantage of R10 and the representation of ReBel is the case where the number of paths is prohibitively large. In this case R7 or some other edge based reduction is likely to be more efficient. As our experiments show this is not the case on the IPC domains tested. In the general case, a meta-reduction heuristic trading off the advantages of different operators would be useful. We discuss our implementation and experimental results in the next sections.

## 4.2 The R11 Reduction

Consider the FODD $B$ in Figure 1(a). Clearly, with no background knowledge this diagram cannot be reduced. Now assume that the background knowledge $\mathcal{B}$ contains a rule $\forall x, [q(x) \rightarrow p(x)]$. In this case if there exists a valuation that reaches the 1 leaf, there must be another such valuation $\zeta$ that agrees on the values of $x$ and $y$. $\zeta$ dominates the other valuations under the $max$ aggregation semantics. The background knowledge rule implies that for $\zeta$, the test at the root node is redundant. However, we cannot set the left child of the root to 0 since the entire diagram will be eliminated. Therefore R7 is not applicable, and similarly none of the other existing reductions is applicable. Yet redundancies like the given example arise often in runs of the value iteration algorithm. This happens naturally, without the artificial background knowledge used for our example but the corresponding diagrams are too large to include in the text. The main reason for such redundancies is that standardizing apart (which was discussed above) introduces multiple renamed copies of the same atoms in the different diagrams. When the diagrams are added, many of the atoms are redundant but some are not removed by old operators. These atoms may end up in a parent-child relation with weak implication from child to parent, similar to the example given. We introduce the R11 reduction operator that can handle such situations. R11 reduces the FODD in Figure 1(a) to the FODD in Figure 1(b).

Let $B$ be a FODD, $n$ a node in $B$, $e$ an edge such that $e \in \{n_{\downarrow t}, n_{\downarrow f}\}$, $e' = \text{sibling}(e)$ (so that when $e = n_{\downarrow t}$, $e' = n_{\downarrow f}$ and vice versa), and $P$ the set of all paths from the root to a non-zero leaf going through edge $e$. Then the reduction $R11(B, n, e)$ drops node $n$ from diagram $B$ and connects its parents to target$(e)$. We need two conditions for the applicability of R11. The first requires that the sibling is a zero valued leaf.

**Condition 1** target$(e') = 0$.

The second requires that valuations that are rerouted by R11 when traversing $B'$, that is valuations that previously reached the 0 leaf and now traverse some path in $P$, are dominated by other valuations giving the same value.

**Condition 2** $\forall p \in P, \mathcal{B} \models [\exists \vec{x^p}, \text{PF}(p) \backslash n^e.lit \wedge n^{e'}.lit] \rightarrow [\exists \vec{x^p}, \text{PF}(p)]$.

The next theorem shows that R11 is sound. The proof shows that by condition 2 the rerouted valuations do not add value to the diagram.

**Theorem 4** *If $B' = R11(B, n, e)$, and conditions 1 and 2 hold, then $\forall$ interpretations $I$, $Map_B(I) = Map_{B'}(I)$.*

*Proof:* Let $I$ be any interpretation and let $Z$ be the set of all valuations. We can divide $Z$ into three disjoint sets depending on the path taken by valuations in $B$ under $I$. $Z^e$ - the set of all valuations crossing edge $e$, $Z^{e'}$ - the set of all valuations crossing edge $e'$ and $Z^{other}$ - the set of valuations not reaching node $n$. We analyze the behavior of the valuations in these sets under $I$.

- Since structurally the only difference between $B$ and $B'$ is that in $B'$ node $n$ is bypassed, all paths from the root to a leaf that do not cross node $n$ remain untouched. Therefore $\forall \zeta \in Z^{other}, Map_B(I, \zeta) = Map_{B'}(I, \zeta)$.

- Since, in $B'$ the parents of node $n$ are connected to target($e$), all valuations crossing edge $e$ and reaching target($e$) in $B$ under $I$ will be unaffected in $B'$ and will, therefore, produce the same map. Thus $\forall \zeta \in Z^e$, $Map_B(I, \zeta) = Map_{B'}(I, \zeta)$.

- Now, let $m$ denote the node target($e$) in $B$. Under $I$, all valuations in $Z^{e'}$ will reach the 0 leaf in $B$ but they will cross node $m$ in $B'$. Depending on the leaf reached after crossing node $m$, the set $Z^{e'}$ can be further divided into 2 disjoint subsets. $Z^{e'}_{zero}$ - the set of valuations reaching a 0 leaf and $Z^{e'}_{nonzero}$ - the set of valuations reaching a non-zero leaf. Clearly $\forall \zeta \in Z^{e'}_{zero}$, $Map_B(I, \zeta) = Map_{B'}(I, \zeta)$.

  By the structure of $B$, every $\zeta \in Z^{e'}_{nonzero}$, traverses some $p \in P$, that is, $(\text{PF}(p) \backslash n^e.lit \wedge n^{e'}.lit)\zeta$ is true in $I$. Condition 2 states that for every such $\zeta$, there is another valuation $\eta$ such that $(\text{PF}(p))\eta$ is true in $I$, so $\eta$ traverses the same path. However, every such valuation $\eta$ must belong to the set $Z^e$ by the definition of the set $Z^e$. In other words, in $B'$ every valuation in $Z^{e'}_{nonzero}$ is dominated by some valuation in $Z^e$.

From the above argument we conclude that in $B'$ under $I$, every valuation either produces the same map as in $B$ or is dominated by some other valuation. Under the max aggregation semantics, therefore, $Map_B(I) = Map_{B'}(I)$. □

## 5. FODD-Planner

In this section we discuss the system FODD-Planner that implements the VI algorithm with FODDs. FODD-Planner employs a number of approximation techniques that yield further speedup. The system also implements extensions of the basic VI algorithm that allow it to handle action costs and universal goals. The following sections describe these details.

### 5.1 Value Approximation

Reductions help keep the diagrams small in size by removing redundancies but when the true $n$ step-to-go value function itself is large, legal reductions cannot help. There are domains where the true value function is unbounded. For example in the tireworld domain from the international planning competition, where the goal is always to get the vehicle to a destination city, one can have a chain of cities linked to one another up to the destination. This chain can be of any length. Therefore when the value function is represented using state abstraction, it must be unbounded. As a result SDP-like algorithms are less effective on domains where the dynamics lead to such transitive structure and every iteration of value iteration increases the size of the $n$ step-to-go value function (Kersting et al., 2004; Sanner & Boutilier, 2009). In other cases the value function is not infinite but is simply too large to manipulate efficiently. When this happens we can resort to approximation keeping as much of the structure of the value function as possible while maintaining efficiency. One must be careful about the tradeoff here. Without approximation the runtime can be prohibitive and too much approximation causes loss of structure and value. We next present three methods to get approximations which act at different levels in the algorithm.

### 5.1.1 Not Standardizing Apart Action Variants

Standardizing apart the diagrams of action variants before adding them is required for the correctness of the FODD based VI algorithm. That is, if we do not standardize apart action variant diagrams before adding them, the value given to some states may be lower than the true value (Wang et al., 2008). Intuitively, this is true since different paths in the value function share atoms and variables. Now, for a fixed action, the best variable binding and corresponding value for different action variants may be different. Thus, if the variables are forced to be the same for the variants, we may rule out viable combinations of value. On the other hand, the value obtained if we do not standardize apart is a lower bound on the true value. This is because every path in the diagram resulting from not standardizing apart is present in the diagram resulting from standardizing apart. Although the value is not exact, not standardizing apart leads to more compact diagrams, and can therefore be useful in speeding up the algorithm. We call this approximation method *non-std-apart* and use it as a heuristic to speed up computation. Although this heuristic may cause loss of structure in the representation of the value function, we have observed that in practice it gives significant speedup while maintaining most of the relevant structure. This approximation is used in some of the experiments described below.

### 5.1.2 Merging Leaves

The use of FODDs also allows us to approximate the value function in a simple and controlled way. Here we follow the approximation techniques of APRICODD (St-Aubin et al., 2000) where they were used for propositional problems. The idea is to reduce the size of the diagram by merging substructures that have similar values. One way of doing this is to reduce the precision of the leaf values. That is, for a given precision value $\epsilon$, we join leaves whose value is within $\epsilon$. This, in turn, leads to reduction of the diagram because subparts of the diagram that previously pointed to different leaves, now point to the same leaf. The granularity of approximation, however, becomes an extra parameter for the system and has to be chosen carefully. Details are provided in the experiments below.

### 5.1.3 Domain Determinization

Previous work on stochastic planning has discovered that for some domains one can get good performance by pretending that the domain is deterministic and re-planning if unexpected outcomes are reached (Yoon et al., 2007). Here we use a similar idea and determinize the domain in the process of policy generation. This saves significant amount of computation and avoids the typical increase in size of the value function encountered in step 2 of the VI algorithm. Domains can be determinized in many ways. We choose to perform determinization by replacing every stochastic action with its most probable deterministic alternative. This is done only once prior to running VI. Although this method of determinization is sub-optimal for many domains, it makes sense for domains where the most probable outcome corresponds to the successful execution of an action (Little & Thibaux, 2007) as is the case in the domains we experimented with. Note that the determinization only applies to the process of policy generation. When the generated policy is deployed to solve planning problems, it does so under the original stochastic environment. This approximation is used in some of the experiments described below.

## 5.2 Extensions of the VI Algorithm

FODD-Planner makes two additional extensions to the basic algorithm. This allows the handling of action costs, arbitrary conjunctive goals as well as universal goals.

### 5.2.1 Handling Action Costs

The standard way to handle action costs is to replace $R(s, a)$ by $R(s, a) - Cost(s, a)$ in the VI algorithm. However, our formalism using FODDs relies on the fact that all the leaves (and thus values) are non-negative. To avoid this difficulty, we note that action costs can be supported as long as there is at least one zero cost action. To see this recall the VI algorithm. The appropriate place to add action costs is just before the Object Maximization step. However, because this step is followed by maximizing over the action diagrams, if at least one action has 0 cost (if not we can create a *no-op* action), the resultant diagram after maximization will never have negative leaves. Therefore we safely convert negative leaves before the maximization step to 0 and thereby avoid conflict with the reduction procedures.

### 5.2.2 Handling Universal Goals

FODDs with *max* aggregation cannot represent universal quantifiers. Therefore our VI algorithm cannot handle universal goals at the abstract level (though see Joshi et al. (2009) for a formalism that does accept arbitrary quantifiers). For a concrete planning problem with a known set of objects we can instantiate the universal goal to get a large conjunctive goal. In principle we can run VI and policy generation for this large conjunctive goal. However, this would mean that we cannot plan off-line to get a generic policy and must replan for each problem instance from scratch. Here we follow an alternative heuristic approach previously introduced by Sanner and Boutilier (2009) and use an approximation of the true value function, that results from a simple additive decomposition of the goal predicates.

Concretely, during off-line planning we plan separately for a generic version of each predicate. For example in the transportation domain discussed above we will plan for the generic predicate box-in-city$(box, city)$ as well as other individual predicates. Then at execution time, when given a concrete goal, we approximate the true value function by the sum of the generic versions over each ground goal predicate. This is clearly not an exact calculation and will not work in every case. On the other hand, it considerably extends the scope of the technique and works well in many situations.

## 5.3 The FODD-Planner System

We implemented the FODD-Planner system, plan execution routines and evaluation routines under Yap Prolog 5.1.2. Our code and domain encodings as used in the experiments reported in the next section are available at `http://code.google.com/p/foddplanner/` under tag `release11-JAIR2011`.

Our implementation uses a simple theorem prover that supports background knowledge by a procedure we call "state flooding". That is, to prove $\mathcal{B} \models X \to Y$, where $X$ is a ground conjunction (represented in Prolog as a list), we "flood" $X$ using rules of the background knowledge using the following simple steps until convergence.

1. Generate $Z$, the set of all ground literals that can be derived from $X$ and the rules of background knowledge.

2. Set $X = X \cup Z$.

When $X$ has converged we test for membership of $Y$ in $X$. Because of our restricted language, the reasoning problem is decidable and our theorem prover is complete.[2]

The overall algorithm is the same as SDP except that all operations are performed on FODDs and reductions are applied to keep all intermediate diagrams compact. In the experiments reported below, we use all previously mentioned reductions (R1 ⋯ R11) except R7-replace. We applied reductions iteratively until no reduction was applicable on the FODD. There is no correct order to apply the reductions in the sense that any reduction when applied can give rise to other reductions. Heuristically we chose an order where we hope to get as much of the diagram reduced as soon as possible. We apply reductions in the following order. We start by applying R10 twice with a different DPO each time. The first DPO is generated by breaking ties in favor of shorter paths. The second is generated by reversing the order of equal valued paths in the first DPO. With R10 we hope to catch many redundant edges early. R10 is followed by R7-drop to remove redundant nodes connected to the edges removed by R10. After this, we apply a round of all strong reductions followed by R9 to remove the redundant equality nodes. R9 is followed by another round of strong reductions. This sequence is performed iteratively until the diagram is stable. In the FODD-PLANNER strong reductions are automatically applied every time two diagrams are combined (using the apply algorithm (Wang et al., 2008)) and weak reductions are applied every time two diagrams are combined except during regression by block combination. We chose to apply R11 only twice in every iteration - once after regression and once just before the next iteration. This setting for application of reduction operators is investigated experimentally and discussed in Section 6.1.

To handle complex goals we use the additive goal decomposition. For each generic goal atom $g$ we run the system for the specified number of iterations, but at the last iteration we do not perform step 4 of the algorithm. This yields a set of functions, $Q^{g,A}$, parameterized by action and generic goal that implicitly represent the policy. To improve on line execution time using this policy we extract the set of paths from the $Q$ functions and perform logical simplification on these paths removing implied atoms and directly applying equalities when they are in the path formula. This is the final form of the policy from the off-line planning phase. for the on-line phase, given a concrete problem state and goal we identify potential actions, and for each action find the top ranking rule for each concrete goal atom $g$. These are combined to give the total value for each action and the action with the highest value is chosen, breaking ties randomly if it is not unique.

---

2. An alternative to the list representation of $X$ would have been to utilize the Prolog database to store the literals of $X$ and employ the Prolog engine to query $Y$. However, in our experience with Yap, it becomes expensive to *assert* (and *retract*) the literals of $X$ to (from) the Prolog database so that the list representation is faster.

## 6. Experimental Results

We ran experiments on a standard benchmark problem as well as probabilistic planning domains from the international planning competitions (IPC) held in 2004, 2006 and 2008. The probabilistic track of the IPC provides domain descriptions in the PPDDL language (Younes, Littman, Weissman, & Asmuth, 2005). We encoded the TVDs and probability and reward functions for these domains by translating the PPDDL manually in a straightforward manner.[3] All experiments were run on a Linux machine with an Intel Pentium D processor running at 3 GHz, with 2 GB of memory. Following IPC standards, all timings, rewards and plan-lengths we report are averages over 30 rounds. For each domain, we constructed by hand background knowledge restricting arguments of predicates (e.g. a box can only be at one city in any time so $Bin(b, c_1), Bin(b, c_2) \rightarrow (c_1 = c_2)$). As discussed above, this is useful in the process of simplifying diagrams.

### 6.1 Merits of Reduction Operators

The following subsections present our main results showing performance in solving planning problems from IPC. Before discussing these we first investigate and illustrate the merits of the various reduction operators in terms of their effect on off-line planning time. The experiments are performed on the tireworld and boxworld domains that are described in more detail below. For this section is suffices to consider the domains as typical cases we might have to address in solving planning problems and focus on the differences between reductions.

In the first set of experiments we compare the run time with R7 and R10 in the context of other reductions. Since R10 and R7 are both edge removal reductions and R7-drop is used in conjunction with both, we compare R10 to R7-replace directly under all configurations of R9 and R11. Except for the choice of reductions used the experimental setup is exactly the same as detailed above. Figures 7 and 8 show the time to build a policy over varying number of iterations for different settings of these weak reduction operators for the boxworld and the tireworld domains. The figures clearly show the superiority of R10 over R7-replace. All combinations with R7-replace have prohibitively large run times at 3 or 4 iterations. With or without R9 and R11, R10 is orders of magnitude more efficient than R7-replace. It is for this reason that in all future experiments we used R10 instead of R7-replace. The experiments also demonstrate that without the new reduction operators presented in this paper the FODD-PLANNER would be too slow to run sufficient iterations of the VI algorithm as done in the following subsections to yield good planning performance. Figure 8 shows that for boxworld R11 hinders R7. It appears that in this case, the application of R11 limits the applicability of R7 causing larger diagrams and thus further slowing down VI.

Figures 9 and 10 show the relative merits of R9 and R11 in the presence of R10 for the two domains. The figures are similar to the previous two plots except that we focus on the relevant portion of the CPU time axis. Clearly R11 is an important reduction and it makes planning more efficient in both settings (just R10 and R10+R9). R9 is less effective in tireworld. In boxworld, however the presence of R9 clearly improves planning efficiency for both settings, and the best performance is achieved in the setting using R10+R9+R11. In addition, R9 targets the removal of equality nodes which no other reduction does directly.

---

3. The FODD formalism cannot capture all of PPDDL. In particular since FODDs cannot represent universal quantification, we cannot handle universal action preconditions. On the other hand FODDs can handle universal action effects. Wang (2007) provides an algorithm and a detailed discussion of translation from PPDDL to FODDs.

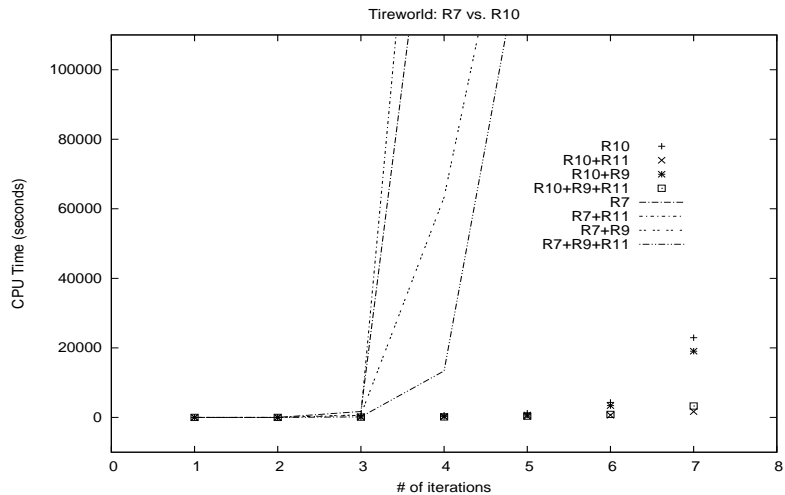

Figure 7: A comparison of planning time taken by various settings of reduction operators over varying number of iterations for tireworld. Four settings of R10 are compared against four settings of R7. All R7 variants do not complete 5 iterations within the time range on the graph and therefore these points are not plotted.

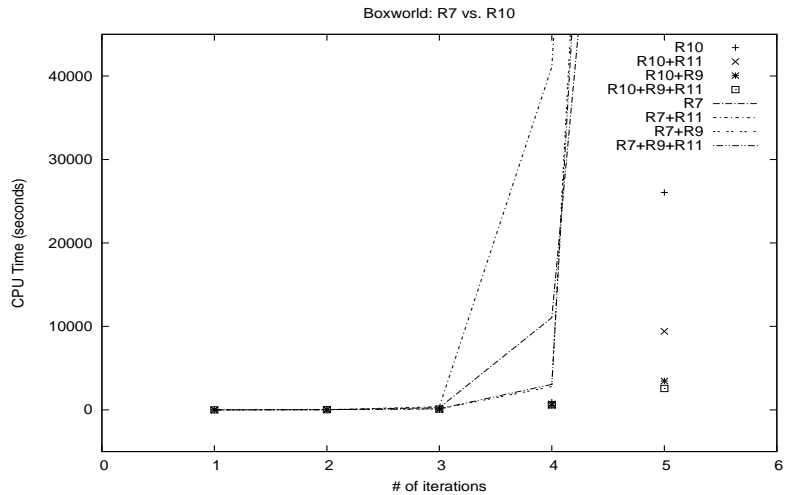

Figure 8: A comparison of planning time taken by various settings of reduction operators over varying number of iterations for boxworld. Four settings of R10 are compared against four settings of R7. All R7 variants do not complete 5 iterations within the time range on the graph and therefore these points are not plotted.

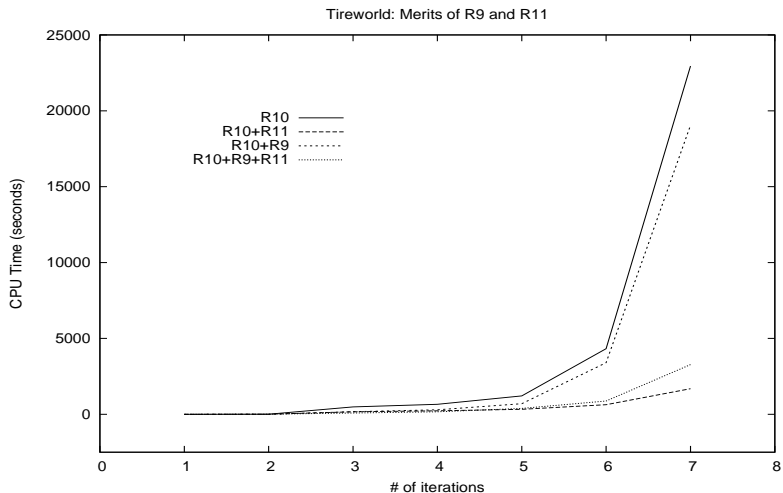

Figure 9: A comparison of the merits of R9 and R11 in the presence of R10 for tireworld.

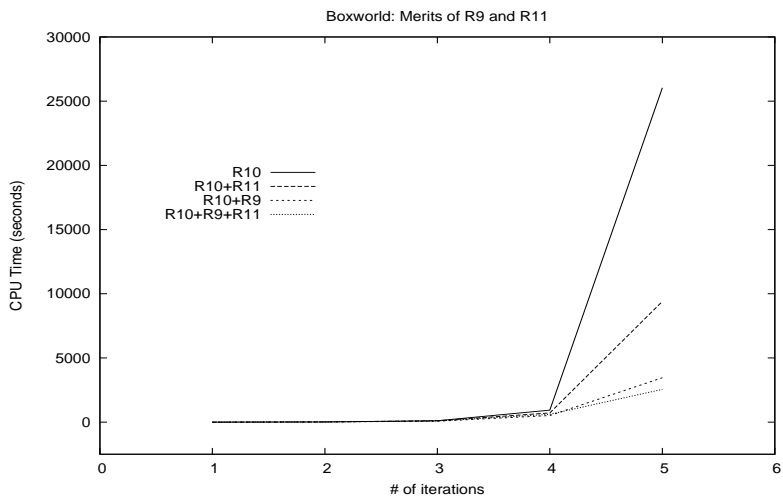

Figure 10: A comparison of the merits of R9 and R11 in the presence of R10 for boxworld.

Based on these results we choose the setting where we employ R10 along with R9 and R11 for the remaining experiments.

## 6.2 The Logistics Benchmark Problem

This is the boxworld problem introduced by Boutilier et al. (2001) that has been used as a standard example for exact solution methods for relational MDPs. The domain consists of boxes, cities and trucks. The objective is to get certain boxes to certain cities by loading, unloading and driving. For the benchmark problem, the goal is the existence of a box in Paris. The load and unload actions are probabilistic and the probability of success of unload depends on whether it is raining or not. In this domain, all cities are reachable from each other. As a result the domain has a compact abstract optimal value function. Note that for this challenge domain there are no concrete planning instances to solve. Instead the goal is to solve the off-line problem and produce the (abstract) optimal solution efficiently. The domain description has 3 predicates of arity 2 and 3 actions each having 2 arguments.

|  | Coverage | Time (ms) | Reward |
|---|---|---|---|
| GPT | 100% | 2220 | 57.66 |
| Policy Iteration with policy language bias | 46.66% | 60466 | 36 |
| Re-Engg NMRDPP | 10% | 290830 | -387.7 |
| FODD-Planner | 100% | 231270 | 70.0 |

Table 1: fileworld domain results

Like ReBel (Kersting et al., 2004) and FOADD (Sanner & Boutilier, 2009) we are able to solve this MDP and identify all relevant partitions of the optimal value function and in fact the value function converges after 10 iterations. FODD-PLANNER performed 10 iterations in under 2 minutes.

## 6.3 The Fileworld Domain

This domain was part of the probabilistic track of IPC-4 (2004) (information on the competitions is accessible at `http://ipc.icaps-conference.org/`). The domain consists of files and folders. Every file obtains a random assignment to a folder at execution time and the goal is to place each file in its assigned folder. There is a cost of 100 to handle a folder and a cost of 1 to place a file in a folder. The optimal policy for this domain is to first get the assignments of files to folders and then handle each folder once, placing all files that were assigned to it. The domain description has 8 predicates of arity 0 to 2 and 16 actions with 0 to 1 arguments.

Results have been published for one problem instance which consisted of thirty files and five folders. Since the goal is conjunctive we used the additive goal decomposition discussed above. We used off-line planning for a generic goal $filed(a)$ and use the policy to solve for any number of files. This domain is ideal for abstract solvers because the optimal value function and policy for a generic goal are compact and can be found quickly. The FODD-PLANNER was able to achieve convergence within 4 iterations even without approximation. Policy generation and execution together took under 4 minutes. Of the 6 systems that competed on this track, results have been published for 3 on the website cited above. Table 1 compares the performance of FODD-PLANNER to the others. We observe that we rank ahead of all in terms of total reward and coverage (both FODD-PLANNER and GPT achieve full coverage).

## 6.4 The Tireworld Domain

This domain was part of the probabilistic track of IPC-5 (2006). The domain consists of a network of locations (or cities). A vehicle starts from one city and moves from city to city with the objective of reaching a destination city. Moves can only be made between cities that are directly connected by a road. In addition, on any move, the vehicle may lose a tire with 40% probability. Some cities have a spare tire that can be loaded onto the vehicle. If the vehicle contains a spare tire, the flat tire can be changed with 50% success probability. This domain is simple but not trivial owing to the possibility of a complex network topology and high probabilities of failure. The IPC description of this domain has 5 predicates of arity 0 to 2 and 3 actions with 0 to 2 arguments.

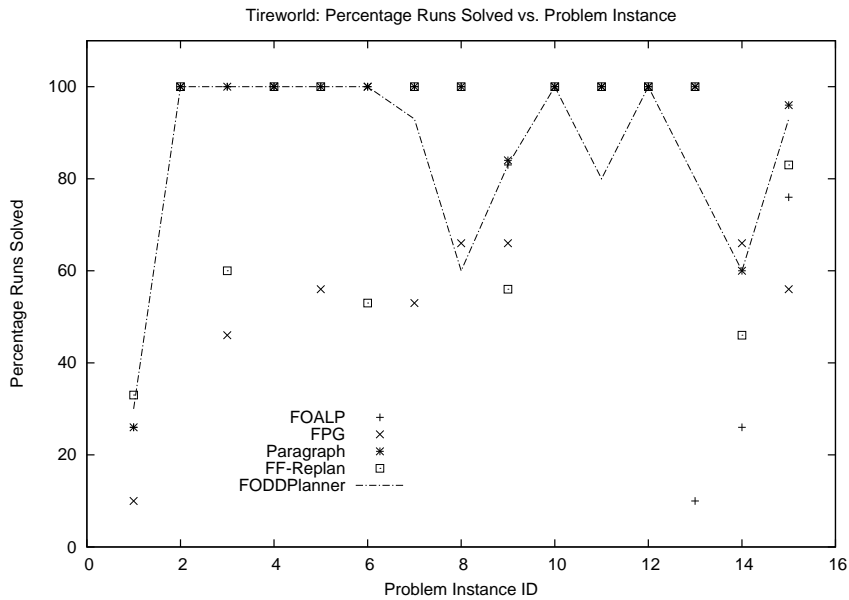

Figure 11: Coverage result of tireworld experiments

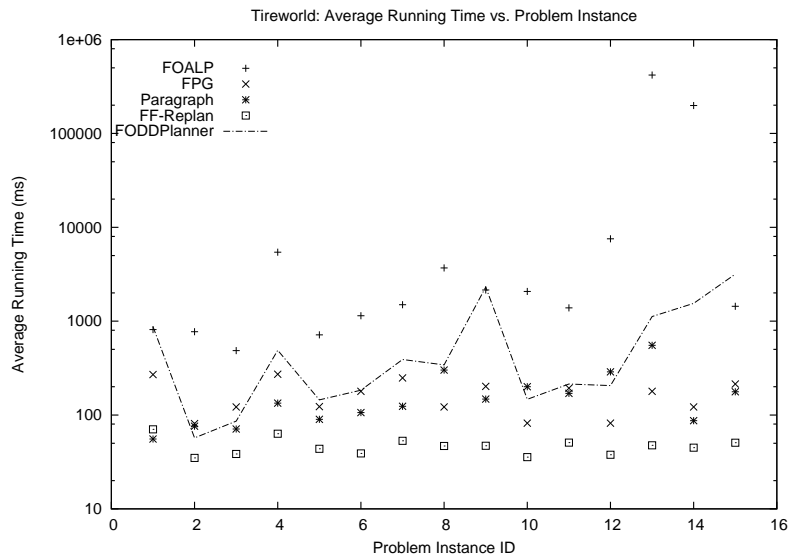

Figure 12: Timing result of tireworld experiments

Participants at IPC-5 competed over 15 problem instances on this domain with varying degree of difficulty. In problem 1 there were 16 locations which were progressively increased by 2 per problem up to 44 locations in problem 15.

To limit off-line planning time we restricted FODD-PLANNER to 7 iterations without any approximation for the first 3 iterations and with the non-std-apart approximation for the remaining iterations. The policy was generated in 55 minutes; this together with the online planning time is within the competition time bound. The performance of FODD-PLANNER and systems competing in the probabilistic track of IPC-5, for which data is published, is summarized in Figures 11, 12, and 13. The figures are indexed by problem instance and show a comparison of the percentage of runs each planner was able to solve (coverage), the average time per instance taken by each planner to generate an online solution, and the average number of actions taken by each planner to reach the goal on every instance. We observe that the overall performance of FODD-PLANNER is competitive with (and in a few cases better than) the other systems. Runtimes to generate online solutions are high for FODD-PLANNER but are comparable to FOALP which is the only other First-Order planner. On the other hand, in comparison with the other systems, we are able to achieve high coverage and short plans on many of the problems.

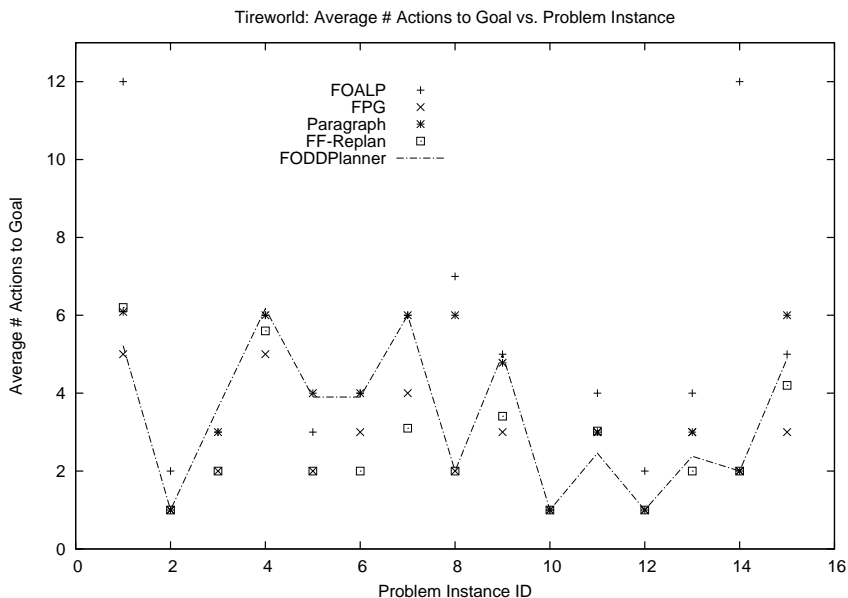

Figure 13: Plan length result of tireworld experiments

## 6.5 Value Approximation by Merging Leaves

Although the tireworld domain can be solved as above within the IPC time limit, one might wish for even faster execution. As we show next, the heuristic of merging leaves provides such a tool, potentially trading off quality of coverage and plan length for faster planning and execution times. Table 2 shows the average reduction in planning time, coverage and planning length achieved when the approximation merging leaves is used. The highest reward obtained in any state is 500. We experimented with reducing precision on the leaves with values between 50.0 and 150.0. As the results demonstrate, for some loss in coverage and planning length, the system can gain in terms of execution time and planning time. For example, with leaf precision of 50.0 (10% of the total value) we get 95.53% reduction in planning time (22 fold speedup) but we lose 15.29% in coverage.[4]

---

4. Note that the measure of plan length, the average over problems solved, is not a good representation of performance when coverage is not full. In this case, if coverage goes down by dropping the harder problems with longer solutions, plan length will appear to be better, but this is clearly not an indication of improved performance.

| Precision | Reduction in Planning Time | Reduction in Execution Time | Reduction in Coverage | Reduction in Plan length |
|---|---|---|---|---|
| 50 | 93.53% | 88.3% | 15.29% | 14.54% |
| 75 | 98.13% | 95.21% | 15.29% | 6.23% |
| 100 | 98.28% | 95.21% | 15.29% | 6.23% |
| 125 | 99.65% | 95.48% | 31.76% | -30.86% |
| 150 | 99.73% | 95.61% | 31.76% | -30.86% |

Table 2: Percentage average reduction in planning time, execution time, coverage and plan length for tireworld under the approximation merging leaves for varying leaf precision values. For example, the first row of the table states that by reducing the precision on the leaves to 50, which is 10% of the largest achievable reward in any state, the planning time was reduced by 93.53% of its original value, average execution time was reduced by 88.3%, average coverage was reduced by 15.29% and average plan length was reduced by 14.54%

### 6.6 Boxworld

In this domain from IPC 2008, the world consists of boxes, trucks, planes and a map of cities. The objective is to get boxes from source cities to destination cities using the trucks and planes. Boxes can be loaded and unloaded from the trucks and planes. Trucks (and planes) can be driven (flown) from one city to another as long as there is a direct road (or air route) from the source to the destination city. The only probabilistic action is *drive*. *drive* works as expected (transporting the truck from the source city to the destination city) with probability 0.8. Occasionally *drive* teleports a truck to the wrong city. The IPC description of this domain includes 11 predicates of arity 2 and 6 actions with 3 arguments.

IPC posted 15 problems with varying levels of difficulty for this domain. In all problems the world consisted of 4 trucks and 2 airplanes. In problems 1 to 3 there were 10 boxes and 5 cities. Problems 4 and 5 had 10 boxes and 10 cities. Problems 6 and 7 had 10 boxes and 15 cities. Problems 8 and 9 had 15 boxes and 10 cities. Problems 10, 11 and 12 had 15 boxes and 15 cities. Competition results show that RFF (Teichteil-Koenigsbuch et al., 2008) was the *only* system that solved any of the 15 problems. Neither RFF nor FODD-PLANNER could solve problems 13 to 15; hence we omit results for those.

To limit off-line planning time we determinized this domain (making *drive* deterministic) and restricted FODD-PLANNER to 5 iterations. Since the domain was determinized, there was only one alternative per action. Therefore the the non-std-apart approximation has no effect here. The policy was generated in 42.6 minutes. The performance of FODD-PLANNER and RFF is summarized in Figures 14, 15, and 16. The figures show a comparison of the percentage of runs each planner was able to solve (coverage), the average reward achieved per problem instance, and the average number of actions taken by each planner to reach the goal on every instance.

As can be seen FODD-PLANNER has lower coverage than RFF. However, our performance is close to RFF in terms of accumulated reward and consistently better in terms of plan length even on problems where we achieve full coverage.

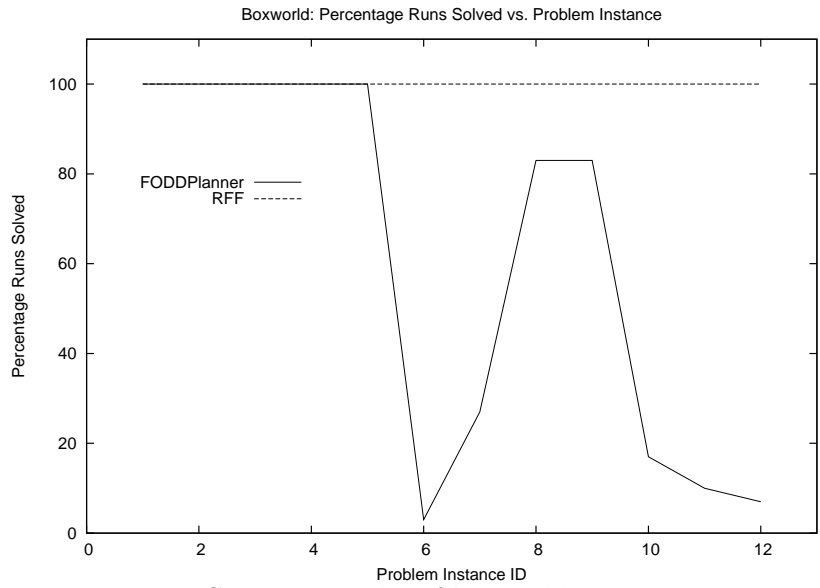

Figure 14: Coverage results of boxworld experiments

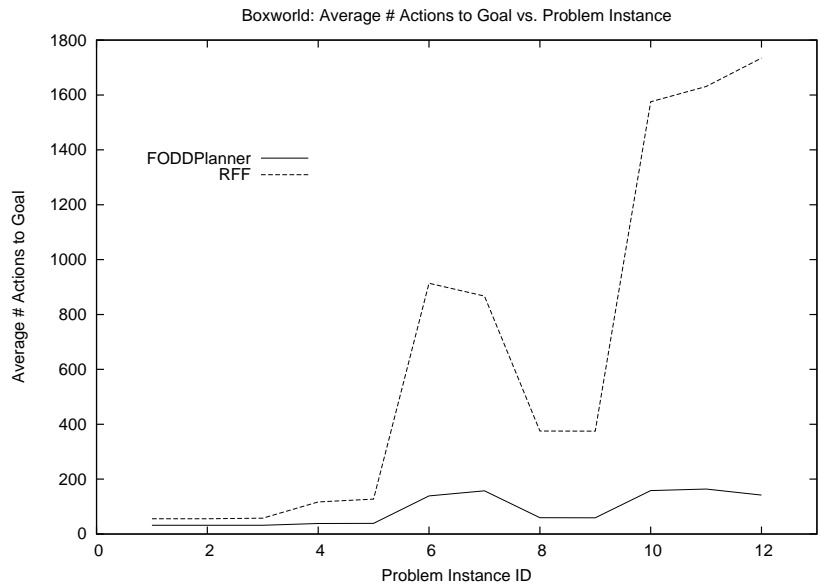

Figure 15: Plan length results of boxworld experiments

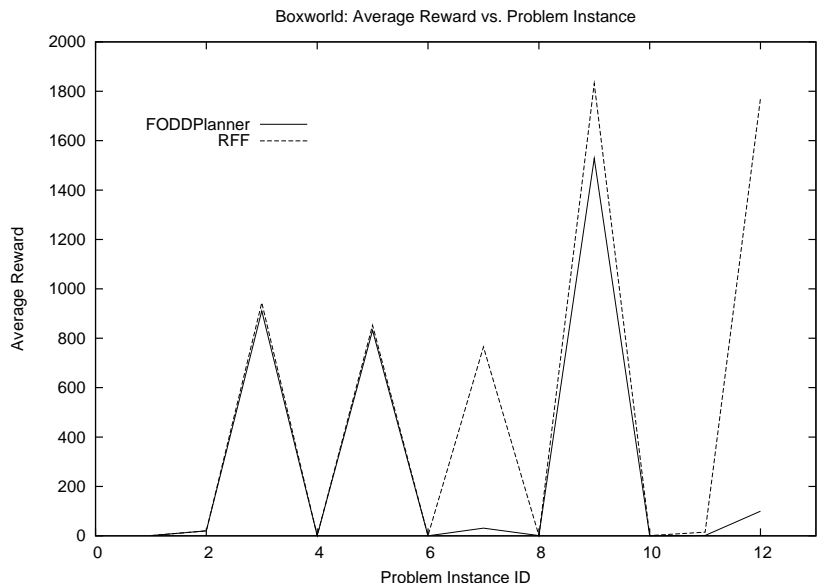

Figure 16: Average reward results of boxworld experiments

In this domain we experienced long plan execution times (10 minutes per round on hard problems and about 15 seconds per round on the easier problems). This points to the complexity of the instances and could be a one reason for the failure of other planning systems at IPC where a strict time bound was observed, and for the failure of RFF on problems 13, 14 and 15. Thus, although the performance of our system is promising, reducing online execution time is crucial. As shown above, for some domains the technique of merging leaves can lead to such improvement at the cost of some reduction in performance. Unfortunately, for this domain merging leaves did not provide any advantage. As in tireworld, there is a clear tradeoff between the quality of coverage and planning time. However the switch is abrupt and to gain significantly in execution time one incurs a significant loss in coverage. Improving the runtime for online application of our policies is an important aspect for future work.

## 7. Related Work

The introduction briefly reviewed previous work on MDPs, propositionally factored MDPs and RMDPs focusing on work that is directly related to the ideas used in this paper. There have been several other solution formalisms for RMDPs that combine dynamic programming with other ideas to yield successful systems. These include approaches that combine dynamic programming with linear function approximation (Sanner & Boutilier, 2009), forward search (Hölldobler et al., 2006) and machine learning (Fern, Yoon, & Givan, 2006; Gretton & Thiebaux, 2004). All of these yielded strong implementations that participated in some planning competitions. Other works do not directly use dynamic programming. For instance Guestrin, Koller, Gearhart, and Kanodia (2003a) present an approach using additive value functions based on object classes and employ linear programming to solve the RMDP. Mausam and Weld (2003) employ SPUDD (Hoey et al., 1999) to solve ground instances of an RMDP, generate training data from the solutions and learn a lifted value function from the training data using a relational tree learner. Gardiol and Kaelbling (2003) apply methods from probabilistic planning to solve RMDPs.

In the most closely related work that preceded our effort, Sanner and Boutilier (2009) developed a relational extension of linear function approximation techniques for factored MDPs. The value function is represented as a weighted sum of basis functions, each denoting a partition of the state space. The difference from the work on factored MDPs is that these basis functions are First-Order formulas and thus the value function is valid for any domain size (this is the same fundamental advantage that RMDP solvers have over ground MDP solvers). They develop methods for automatic generation of First-Order constraints in a linear program and automatic generation of basis functions that show promise in solving some domains from the IPC. The work of Sanner and Boutilier is thus an extension of the work on linear representations for propositionally factored MDPs (e.g., Guestrin et al., 2003b) to capture relational structure. In a similar view the work on FODD-PLANNER is a relational extension of the work on ADD based solvers for propositionally factored MDPs (Hoey et al., 1999). In this context it is interesting to note that Sanner and Boutilier also developed a relational extension of ADDs they call FOADDs. In contrast with FODDs, nodes in FOADDs are labeled with closed First-Order formulas.[5] Sanner and Boutilier report on an implementation that was able to provide exact solutions for simple problems, but they developed and applied the approach using linear function approximation for more complex problems. Our experiments do use approximation and they demonstrate that FODDs can be used to solve problems at least of the complexity currently employed in the IPC.

Another important body of work is pursued by Relational Reinforcement Learning (RRL) (Tadepalli, Givan, & Driessens, 2004) where techniques from reinforcement learning are used to learn or construct value functions and policies for relational domains. RRL followed from the seminal work of Dzeroski, De Raedt, and Driessens (2001) whose algorithm involved generating state-value pairs by state space exploration (biased in favor of state-action pairs with high estimated value) and learning a relational value function tree from the collected data. In a sense the First-Order decision trees used by Dzeroski et al. (2001) are similar to FODDs. However, there is an important difference in the semantics of these representations with strong implications for computational properties. While the trees employ semantics based on traversal of a single path, FODD semantics are based on aggregating values generated by traversal of multiple paths. We have previously argued (Wang et al., 2008) that the FODD semantics are much better suited for dynamic programming solutions. There have been several approaches to RRL in recent years showing nice performance (for example, Driessens & Dzeroski, 2004; Kersting & De Raedt, 2004; Walker,

---

5. As discussed by Sanner and Boutilier (2009) it is hard to characterize the exact relationship between FOADDs and FODDs in terms of representation and computational properties. An anonymous reviewer kindly provided the following example that shows that in some cases FODDs might be more compact than FOADDs. Consider a domain with $n$ unary predicates $A_1(\cdot), \ldots, A_n(\cdot)$ capturing some object properties and consider the formula $\exists x, A_1(x) \text{ Xor } A_2(x) \text{ Xor } \ldots \text{ Xor } A_n(x)$ where $n$ is odd. The formula requires that there exists an object for which an odd number of properties $A_i(\cdot)$ hold. Due to their restriction to use only the connectives And, Or and Not, the FOADDs must rewrite this formula in a way that yields a representation (for example in its DNF form) whose size is exponential in the number of predicates. On the other hand, one can represent this formula with a linear size FODD, similar to the representation of parity functions with propositional BDDs.

Torrey, Shavlik, & Maclin, 2007; Croonenborghs, Ramon, Blockeel, & Bruynooghe, 2007) although they are applied to problems of smaller scale than the ones from the IPC. An excellent overview of the various solutions methods for RMDPs is provided by van Otterlo (2008).

## 8. Conclusion and Future Work

The main contribution of this paper is the introduction of FODD-Planner, a relational planning system based on First Order Decision Diagrams. This is the first planning system that uses lifted algebraic decision diagrams as its representation language and successfully solves planning problems from the international planning competition. FODD-Planner provides several improvements over previous work on FODDs (Wang et al., 2008). The improvements include the reduction operators R10, R11 the Sub-apart operator, and several speedup and value approximation techniques. Taken together, these improvements provide substantial speedup making the approach practical. Therefore, the results show that abstraction through compact representation is a promising approach to stochastic planning.

Our work raises many questions concerning foundations for FODDs and their application to solve RMDPs. The first is the question of reductions. Our set of reductions is still heuristic and does not guarantee a canonical form for diagrams which is instrumental for efficiency of propositional algorithms. Identifying such "complete" sets of reductions operators and canonical forms is an interesting challenge. Identifying a practically good set of operators trading off complexity for reduction power is crucial for further applicability. In recent work (Joshi, Kersting, & Khardon, 2010) we developed practical variants of model-checking reductions (Joshi et al., 2009) demonstrating significant speedup over the system presented here. Another improvement may be possible by using the FODD based policy iteration algorithm (Wang & Khardon, 2007). This may allow us to avoid approximation of infinite size value functions in cases where the policy is still compact. Another direction is the use of the more expressive GFODDs (Joshi et al., 2009) that can handle arbitrary quantification and can therefore be applied more widely. Finally this work suggests the potential of using FODDs as the underlying representation for relational reinforcement learning. Therefore, it will be interesting to develop learning algorithms for FODDs.

## Acknowledgments

This work was partly supported by NSF grants IIS 0936687 and IIS 0964457. Saket Joshi was additionally supported by a Computing Innovation Postdoctoral Fellowship. Some of the experiments reported in this paper were performed on the Tufts Linux Research Cluster supported by Tufts UIT Research Computing. We thank Kristian Kersting for valuable input on the system and insightful discussions.

## References

Bahar, R., Frohm, E., Gaona, C., Hachtel, G., Macii, E., Pardo, A., & Somenzi, F. (1993). Algebraic decision diagrams and their applications. In *IEEE /ACM ICCAD*, pp. 188–191.

Bellman, R. (1957). *Dynamic Programming*. Princeton University Press, Princeton, NJ.

Blum, A., & Furst, M. (1997). Fast planning through planning graph analysis. *Artificial Intelligence, 90(1-2)*, 279–298.

Blum, A., & Langford, J. (1998). Probabilistic planning in the graphplan framework. In *Proceedings of the Fifth European Conference on Planning*, pp. 8–12.

Bonet, B., & Geffner, H. (2001). Planning as heuristic search. *Artificial Intelligence, 129*, 5–33.

Boutilier, C., Dean, T., & Hanks, S. (1999a). Decision-theoretic planning: Structural assumptions and computational leverage. *Journal of Artificial Intelligence Research, 11*, 1–94.

Boutilier, C., Dearden, R., & Goldszmidt, M. (1999b). Stochastic dynamic programming with factored representations. *Artificial Intelligence, 121*, 49–107.

Boutilier, C., Reiter, R., & Price, B. (2001). Symbolic dynamic programming for First-Order MDPs. In *Proceedings of the International Joint Conference of Artificial Intelligence*, pp. 690–700.

Croonenborghs, T., Ramon, J., Blockeel, H., & Bruynooghe, M. (2007). Online learning and exploiting relational models in reinforcement learning. In *Proceedings of the International Joint Conference of Artificial Intelligence*, pp. 726–731.

Driessens, K., & Dzeroski, S. (2004). Integrating guidance into relational reinforcement learning. *Machine Learning, 57*, 271–304.

Dzeroski, S., De Raedt, L., & Driessens, K. (2001). Relational reinforcement learning. *Machine Learning, 43*, 7–52.

Fern, A., Yoon, S., & Givan, R. (2006). Approximate policy iteration with a policy language bias. *Journal of Artificial Intelligence Research, 25(1)*, 75–118.

Fikes, R., & Nilsson, N. (1971). STRIPS: A new approach to the application of theorem proving to problem solving. *Artificial Intelligence, 2(3-4)*, 189–208.

Gardiol, N., & Kaelbling, L. (2003). Envelope-based planning in relational MDPs. In *Proceedings of the International Conference on Neural Information Processing Systems*, pp. 1040–1046.

Gretton, C., & Thiebaux, S. (2004). Exploiting First-Order regression in inductive policy selection. In *Proceedings of the Workshop on Uncertainty in Artificial Intelligence*.

Groote, J., & Tveretina, O. (2003). Binary decision diagrams for First-Order predicate logic. *Journal of Logic and Algebraic Programming, 57*, 1–22.

Guestrin, C., Koller, D., Gearhart, C., & Kanodia, N. (2003a). Generalizing plans to new environments in relational MDPs. In *Proceedings of the International Joint Conference of Artificial Intelligence*, pp. 1003–1010.

Guestrin, C., Koller, D., Parr, R., & Venkataraman, S. (2003b). Efficient solution algorithms for factored MDPs. *Journal of Artificial Intelligence Research, 19*, 399–468.

Hoey, J., St-Aubin, R., Hu, A., & Boutilier, C. (1999). SPUDD: Stochastic planning using decision diagrams. In *Proceedings of the Workshop on Uncertainty in Artificial Intelligence*, pp. 279–288.

Hölldobler, S., Karabaev, E., & Skvortsova, O. (2006). FluCaP: a heuristic search planner for First-Order MDPs. *Journal of Artificial Intelligence Research, 27*, 419–439.

Howard, R. (1960). *Dynamic Programming and Markov Processes*. MIT Press.

Joshi, S., Kersting, K., & Khardon, R. (2009). Generalized First-Order decision diagrams for First-Order Markov decision processes. In *Proceedings of the International Joint Conference of Artificial Intelligence*, pp. 1916–1921.

Joshi, S., Kersting, K., & Khardon, R. (2010). Self-Taught decision theoretic planning with First-Order decision diagrams. In *Proceedings of the International Conference on Automated Planning and Scheduling*, pp. 89–96.

Kautz, H., & Selman, B. (1996). Pushing the envelope: Planning, propositional logic, and stochastic search. In *Proceedings of the National Conference of the American Association for Artificial Intelligence*, pp. 1194–1201.

Kearns, M., & Koller, D. (1999). Efficient reinforcement learning in factored MDPs. In *Proceedings of the International Joint Conference of Artificial Intelligence*, pp. 740–747.

Kersting, K., & De Raedt, L. (2004). Logical Markov decision programs and the convergence of logical TD($\lambda$). In *Proceedings of Inductive Logic Programming*, pp. 180–197.

Kersting, K., van Otterlo, M., & De Raedt, L. (2004). Bellman goes relational. In *Proceedings of the International Conference on Machine Learning*, pp. 465–472.

Khardon, R. (1999). Learning function free Horn expressions. *Machine Learning, 37*(3), 249–275.

Little, I., & Thibaux, S. (2007). Probabilistic planning vs. replanning. In *Proceedings of the ICAPS Workshop on IPC: Past, Present and Future*.

Lloyd, J. (1987). *Foundations of Logic Programming*. Springer Verlag. Second Edition.

Majercik, S., & Littman, M. (2003). Contingent planning under uncertainty via stochastic satisfiability. *Artificial Intelligence, 147*(1-2), 119–162.

Mausam, & Weld, D. (2003). Solving relational MDPs with First-Order machine learning. In *Proceedings of the ICAPS Workshop on Planning under Uncertainty and Incomplete Information*.

Penberthy, J., & Weld, D. (1992). UCPOP: A sound, complete, partial order planner for ADL. In *Principles of Knowledge Representation and Reasoning*, pp. 103–114.

Puterman, M. L. (1994). *Markov decision processes: Discrete stochastic dynamic programming*. Wiley.

Sanner, S., & Boutilier, C. (2009). Practical solution techniques for First-Order MDPs. *Artificial Intelligence, 173*, 748–788.

St-Aubin, R., Hoey, J., & Boutilier, C. (2000). APRICODD: Approximate policy construction using decision diagrams. In *Proceedings of the International Conference on Neural Information Processing Systems*, pp. 1089–1095.

Tadepalli, P., Givan, R., & Driessens, K. (2004). Relational reinforcement learning: An overview. In *Proceedings of the International Conference on Machine Learning '04 Workshop on Relational Reinforcement Learning*.

Teichteil-Koenigsbuch, F., Infantes, G., & Kuter, U. (2008). RFF: A robust FF-based MDP planning algorithm for generating policies with low probability of failure. In *Sixth IPC at ICAPS*.

van Otterlo, M. (2008). *The logic of Adaptive behavior: Knowledge representation and algorithms for adaptive sequential decision making under uncertainty in First-Order and relational domains*. IOS Press.

Walker, T., Torrey, L., Shavlik, J., & Maclin, R. (2007). Building relational world models for reinforcement learning. In *Proceedings of Inductive Logic Programming*, pp. 280–291.

Wang, C. (2007). *First-Order Markov decision processes*. Ph.D. thesis, Tufts University.

Wang, C., Joshi, S., & Khardon, R. (2008). First-Order decision diagrams for relational MDPs. *Journal of Artificial Intelligence Research*, *31*, 431–472.

Wang, C., & Khardon, R. (2007). Policy iteration for relational MDPs. In *Proceedings of the Workshop on Uncertainty in Artificial Intelligence*, pp. 408–415.

Weld, D., Anderson, C., & Smith, D. (1998). Extending graphplan to handle uncertainty and sensing actions. In *Proceedings of the National Conference on Artificial Intelligence*.

Yoon, S., Fern, A., & Givan, R. (2007). FF-Replan: A baseline for probabilistic planning. In *Proceedings of the International Conference on Automated Planning and Scheduling*, pp. 352–359.

Younes, H., Littman, M., Weissman, D., & Asmuth, J. (2005). The first probabilistic track of the international planning competition. *Journal of Artificial Intelligence Research*, *24*(1), 851–887.